\documentclass[sigconf]{acmart}

\usepackage{multirow}
\usepackage{tabu}
\usepackage{fancyhdr}
\AtBeginDocument{%
  \providecommand\BibTeX{{%
    \normalfont B\kern-0.5em{\scshape i\kern-0.25em b}\kern-0.8em\TeX}}}

\copyrightyear{2021}
\acmYear{2021}
\setcopyright{acmcopyright}\acmConference[MM '21]{Proceedings of the 29th ACM International Conference on Multimedia}{October 20--24, 2021}{Virtual Event, China}
\acmBooktitle{Proceedings of the 29th ACM International Conference on Multimedia (MM '21), October 20--24, 2021, Virtual Event, China}
\acmPrice{15.00}
\acmDOI{10.1145/3474085.3475216}
\acmISBN{978-1-4503-8651-7/21/10}

\begin{document}
\fancyhead{}

\title{Few-Shot Fine-Grained Action Recognition via Bidirectional Attention and Contrastive Meta-Learning}

\author{Jiahao Wang$^{1}$,\quad Yunhong Wang$^1$,\quad Sheng Liu$^{1}$,\quad Annan Li$^{1}$*}

\makeatletter
\def\authornotetext#1{
\if@ACM@anonymous\else
    \g@addto@macro\@authornotes{
    \stepcounter{footnote}\footnotetext{#1}}
\fi}
\makeatother
\authornotetext{Corresponding author.}

\affiliation{
 \institution{\textsuperscript{\rm 1}State Key Laboratory of Virtual Reality Technology and System, Beihang University, Beijing} 
 \country{China}
 }
\email{{jhwang, yhwang, liu_sheng, liannan}@buaa.edu.cn}

\def\authors{Jiahao Wang, Yunhong Wang, Sheng Liu, Annan Li}

\renewcommand{\shortauthors}{Jiahao Wang et al.}

\begin{abstract}
Fine-grained action recognition is attracting increasing attention due to the emerging demand of specific action understanding in real-world applications, whereas the data of rare fine-grained categories is very limited.
Therefore, we propose the few-shot fine-grained action recognition problem, aiming to recognize novel fine-grained actions with only few samples given for each class.
Although progress has been made in coarse-grained actions, existing few-shot recognition methods encounter two issues handling fine-grained actions: the inability to capture subtle action details and the inadequacy in learning from data with low inter-class variance. 
To tackle the first issue, a human vision inspired bidirectional attention module (BAM) is proposed. Combining top-down task-driven signals with bottom-up salient stimuli, BAM captures subtle action details by accurately highlighting informative spatio-temporal regions.
To address the second issue, we introduce contrastive meta-learning (CML). Compared with the widely adopted ProtoNet-based method, CML generates more discriminative video representations for low inter-class variance data, since it makes full use of potential contrastive pairs in each training episode.  
Furthermore, to fairly compare different models, we establish specific benchmark protocols on two large-scale fine-grained action recognition datasets. 
Extensive experiments show that our method consistently achieves state-of-the-art performance across evaluated tasks.   
\end{abstract}

\vspace{-2mm}

\vspace{-2mm}

\maketitle

\begin{figure}[t]
  \centering
  \includegraphics[width=\linewidth]{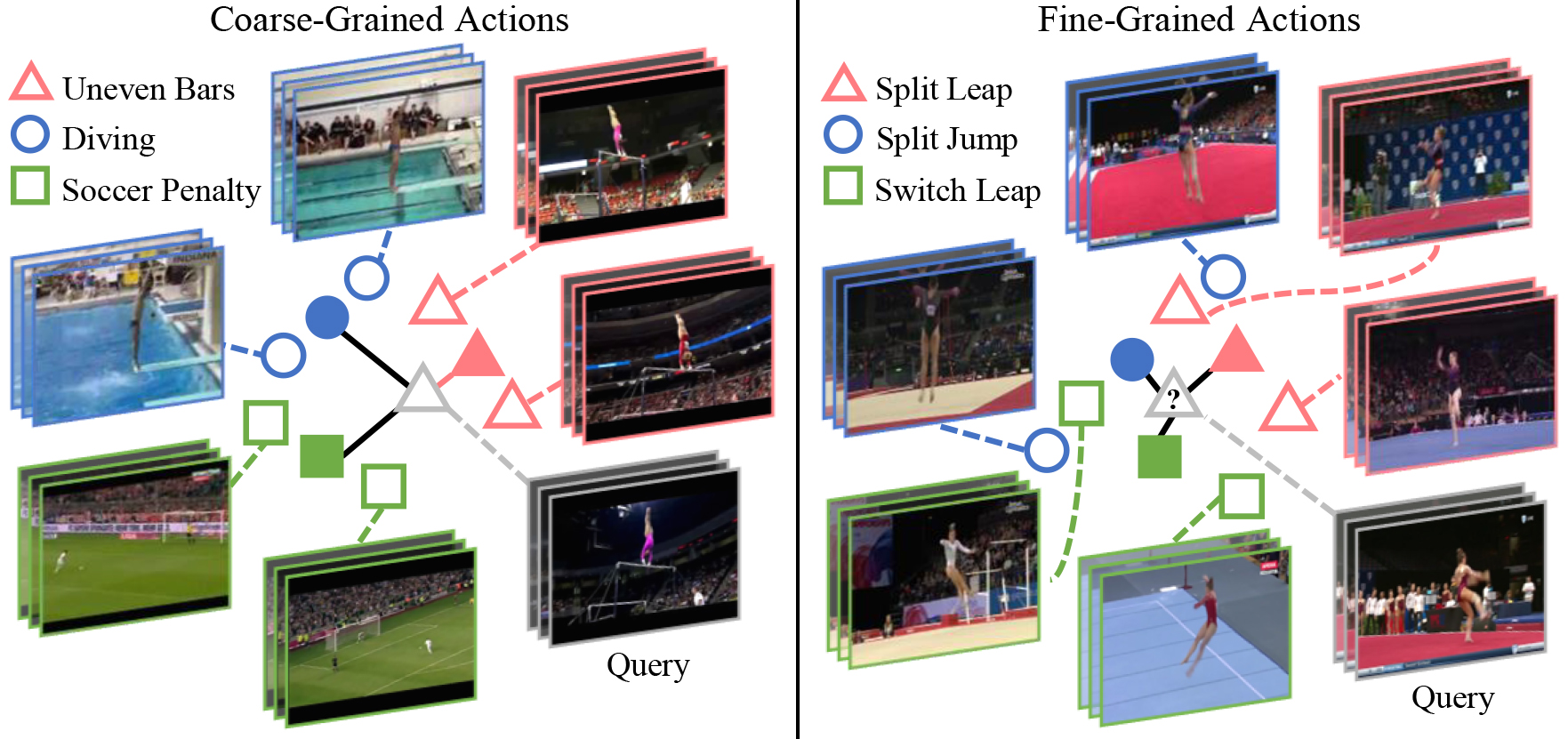}
  \vspace{-5mm}
  \caption{Examples of 3-way 2-shot 1-query action recognition tasks on coarse-grained (left) and fine-grained (right) actions. Different polygons illustrate different categories in the representation space, with solid ones showing the averaged prototypes. Coarse-grained actions can be roughly recognized by the backgrounds, while fine-grained actions differ in subtle motions. Due to the low inter-class variance, averaged prototypes are not discriminative enough for fine-grained action classification.
  } 
  \vspace{-4mm}
  \label{fig:intro}
\end{figure}

\vspace{-1mm}
\section{Introduction}

The demand for fine-grained action recognition is increasing rapidly in real-world applications like sports analytics~\cite{qi2019sports,qi2019stagnet,kong2019joint} and video understanding~\cite{qi2020stc,qi2021semantics,wang2019atrous}. As shown in Figure~\ref{fig:intro}, instead of recognizing coarse-grained actions with apparent difference (e.g. "diving" vs. "gymnastics"), many real-world scenarios demand recognizing actions at finer granularities (e.g. different movements in gymnastics). However, the data of rare fine-grained categories is usually very limited. Take the recently proposed FineGym~\cite{shao2020finegym} dataset as an example, it contains 288 categories of fine-grained gymnastic actions, a large number of which only have less than 10 samples. We thus propose the few-shot fine-grained action recognition problem, aiming to recognize novel fine-grained actions with only few support samples given under each class. 
 
With extensive efforts devoted to few-shot learning (FSL) recently, great success has been achieved in few-shot image classification~\cite{vinyals2016matching,snell2017prototypical,sung2018learning,he2020memory,zou2020compositional}. Spurred by that, attempts are being made to extend FSL to action recognition domain~\cite{zhu2018compound,cao2020few,zhang2020few,qi2020few,fu2020depth,zhu2021few,perrett2021temporal}. Most methods follow the established practice of meta-learning paradigm~\cite{vinyals2016matching}, where models are trained with randomly sampled support and query sets episodically. In each training episode, the classification loss is computed by testing query samples with a classifier built on support samples. To handle multi-shot learning tasks, the prototypical network (ProtoNet)~\cite{snell2017prototypical} is widely adopted. Representations of support samples in each class are simply averaged as prototypes for classification.

Although existing FSL methods have achieved promising results on coarse-grained actions, there are two major issues when extending these methods to fine-grained action recognition. For a clear comparison, we illustrate examples of few-shot recognition tasks on coarse-grained (UCF101~\cite{soomro2012ucf101}) and fine-grained (FineGym~\cite{shao2020finegym}) actions in Figure~\ref{fig:intro}. Firstly, unlike coarse-grained actions, which can be roughly recognized by the background context~\cite{soomro2012ucf101,carreira2017quo}, fine-grained action recognition requires differentiating between subtle details of fine-grained categories~\cite{shao2020finegym}. Such details usually relate to the motion of salient persons/objects, which cannot be explicitly captured by existing few-shot action recognition methods. Secondly, due to the low inter-class variance of fine-grained actions, the averaged representations generated by ProtoNet may not be as discriminative as in coarse-grained actions. Previous works~\cite{snell2017prototypical,sung2018learning} have shown that learning better prototypes requires a larger episodic training batch. While expanding the batch size is prohibitively expensive for video data.

\begin{figure*}[t]
\begin{center}
   \includegraphics[width=0.95\textwidth]{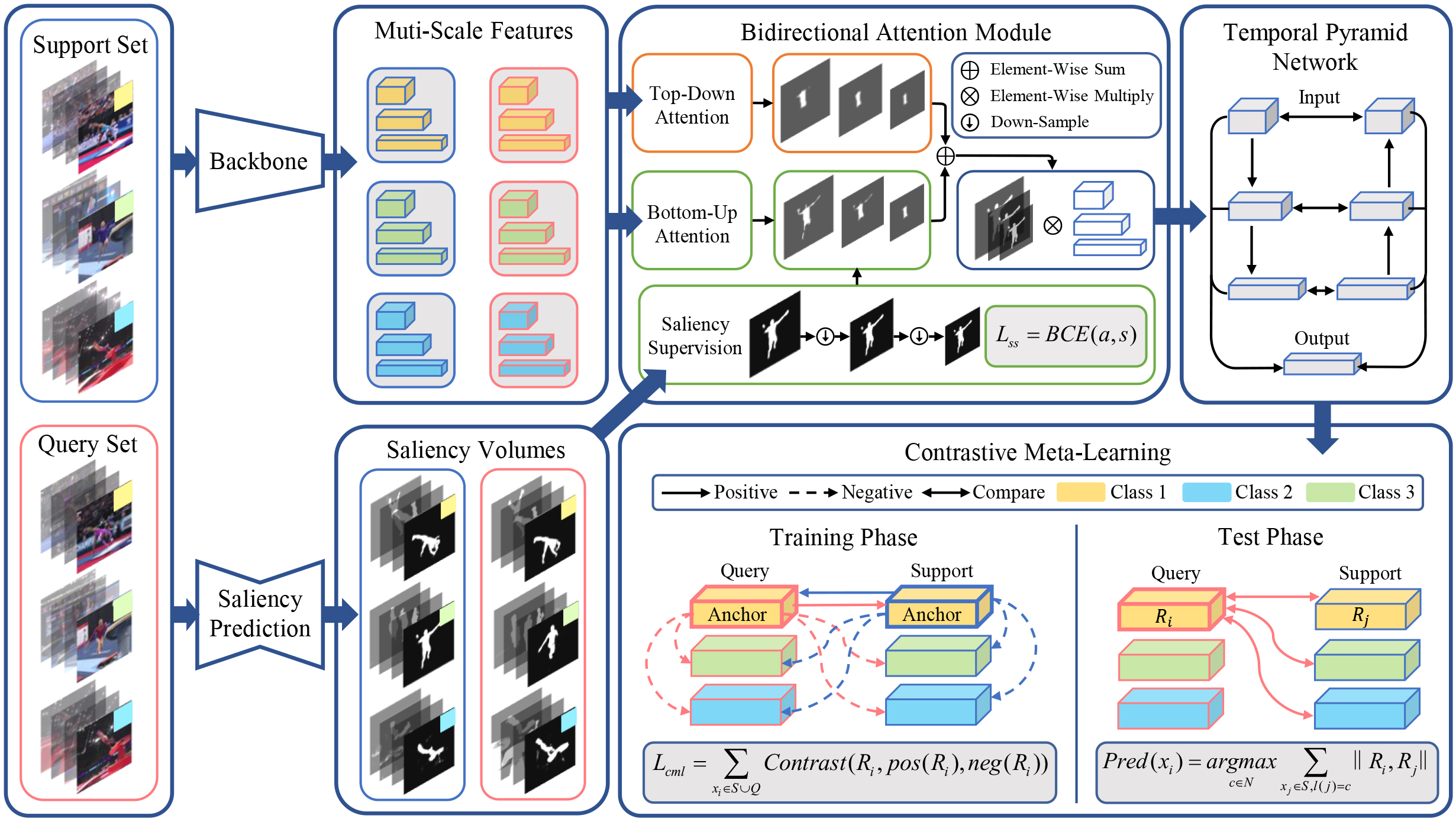}
\end{center}
   \vspace{-4mm}
   \caption{Illustration of proposed framework on a 3-way 1-shot 1-query fine-grained action recognition task. We first extract multi-scale spatio-temporal features from each video clip with a backbone network. Extracted features are then refined by the fusion of top-down and bottom-up attention in the bidirectional attention module. Predicted saliency volumes are down-sampled as ground truth in the saliency supervision loss $L_{ss}$. We utilize a temporal pyramid network to fuse the attended multi-scale features to a unified representation, on which the contrastive meta-learning is performed. During training, Both support and query samples are taken as anchors in the contrastive loss $L_{cml}$. In the test phase, few-shot prediction $Pred(x_i)$ is made by comparing each query sample with every support sample in distance similarity.}
   \vspace{-4mm}
\label{fig:framework}
\end{figure*}

Therefore, we propose a dedicated framework for few-shot fine-grained action recognition addressing the aforementioned issues. As presented in Figure~\ref{fig:framework}, given input videos from the support and query sets of a training episode, we first extract multi-scale spatio-temporal features with a CNN-based backbone network. To address the first issue, a bidirectional attention module (BAM) is introduced to capture sutle action details from extracted features. BAM imitates the top-down and bottom-up attention mechanism of human vision~\cite{buschman2007top,baluch2011mechanisms}. The top-down attention is directly driven by few-shot action recognition task, capturing specific action related features like sports equipments and critical body parts. While the bottom-up attention is supervised by predicted saliency volumes, focusing on class-agnostic saliency features like athletes and objects in motion.
Combining the two together, BAM accurately highlights spatio-temporal regions containing informative action details. 

To tackle the second issue, we propose contrastive meta-learning (CML),  which generates more discriminative video representations for low inter-class variance data compared with ProtoNet. The attended multi-scale features from BAM are first fused by a temporal pyramid network~\cite{yang2020temporal}, generating a unified representation for each video. CML is then performed in this representation space. 
In the training phase, both support and query samples are taken as anchors to construct contrastive pairs. For each anchor, we consider samples with the same label as positives and the others as negatives. In the test phase, more accurate prediction is made by comparing query samples with every support sample rather than simply an averaged prototype. CML makes full use of potential contrastive pairs in each episode, thus unleashing much more contrastive power from limited training samples than the ProtoNet-based method.

Since training and test protocols are quite influential on the result of few-shot recognition problems~\cite{sung2018learning,zhang2020few}, 
we establish the first benchmark for few-shot fine-grained action recognition on two recently proposed large-scale fine-grained action datasets, i.e. FineGym~\cite{shao2020finegym} and HAA500~\cite{chung2020haa500}. We evaluate our method and various challenging competitors under specific protocols and make them available to facilitate future works. Ablative experiments further validate the efficacy of our framework components.

To summarize, our key contributions include:
\vspace{-1mm}
\begin{itemize}
    \item We propose the few-shot fine-grained action recognition problem, which is spawned from real-world demands. A novel framework is devised to accurately recognize unseen fine-grained actions with few support samples.
    \item We combine task-driven and saliency-supervised signals to capture subtle action details, simulating the top-down and bottom-up attention mechanism of human vision.
    \item To generate discriminative representations for low inter-class variance data, we propose contrastive meta-learning, making full use of potential contrastive pairs.
    \item Specific benchmark protocols are established for few-shot fine-grained action recognition, where our method shows significant advantages over the competitors.
\end{itemize}

\vspace{-2mm}
\section{Related Work}

\noindent\textbf{Fine-grained action recognition}. 
Fine-grained action recognition aims to distinguish between sub-classes below basic human action taxonomy~\cite{sun2017taichi,shao2020finegym,chung2020haa500}. Compared with coarse-grained actions, the inter-class variance of fine-grained actions is relatively low, which requires differentiating very subtle motion details. Efforts have been made towards building ideal datasets for fine-grained action recognition over the years~\cite{rohrbach2016recognizing,sun2017taichi,piergiovanni2018fine,shao2020finegym,chung2020haa500,yan2017fine}. Fine-grained actions in individual sports like taichi~\cite{sun2017taichi}, baseball~\cite{piergiovanni2018fine} are collected at first. Recently, large-scale datasets with a great number of action classes like FineGym~\cite{shao2020finegym} and HAA500~\cite{chung2020haa500} are proposed. 

In terms of recognition methods, Piegiovanni and Ryoo~\cite{piergiovanni2018fine} combine CNN models with various feature aggregation methods and find the sub-events scheme is optimal. Zhu et al.~\cite{zhu2018fine} propose a redundancy reduction attention (RRA) module to focus on discriminative feature channels. A recent study on FineGym dataset~\cite{shao2020finegym} shows that existing coarse-grained action recognition models~\cite{wang2016temporal,carreira2017quo} are insufficient to capture subtle motion details. Moreover, traditional recognition methods suffer from issues like overfitting due to scarceness of data in rare fine-grained action categories. 

\noindent\textbf{Few-shot learning.} The goal of few-shot learning (FSL) is to effectively implement learning algorithms with limited data. Existing FSL methods can be roughly categorized into optimization-based~\cite{finn2017model,li2017meta} and metric-based~\cite{vinyals2016matching,sung2018learning,snell2017prototypical, he2020memory, wang2020cooperative} methods. Most state-of-the-arts adopt the episodic meta-learning strategy, which trains the model with randomly sampled support and query sets in each episode. Spurred by the success of FSL in image classification, attempts are being made to explore FSL in action recognition
~\cite{zhu2018compound,cao2020few,zhang2020few,qi2020few,fu2020depth,zhu2021few,perrett2021temporal,zhang2020few-shot}. 
CMN~\cite{zhu2018compound} utilizes memory networks to compress video information into a fixed matrix, which facilitates few-shot recognition by feature matching. OTAM~\cite{cao2020few} computes distance matrices between query and support videos with temporal alignment. Fu et al.~\cite{fu2020depth} leverage depth features to provide additional information for video representation learning. Perrett et al.~\cite{perrett2021temporal} propose the TRX model, which utilizes temporally-corresponding frame tuples to represent action similarity. Nevertheless, none of these methods is designed to handle fine-grained actions. Also, most existing methods adopt ProtoNet~\cite{snell2017prototypical} for meta-learning, which computes averaged prototypes in multi-shot scenarios. 

In comparison, our framework is different in both architecture and meta-learning method. We capture fine-grained action details with bidirectional attention from multi-scale features. The proposed contrastive meta-learning directly contrasts between training samples without creating any prototype representations.

\noindent\textbf{Visual attention mechanism.} 
Neuroscience studies~\cite{ungerleider2000mechanisms,buschman2007top,baluch2011mechanisms} have revealed that human cortex can be focused volitionally by combining top-down signals derived from task demands and bottom-up signals from salient stimuli. Attempts were made to implement such attention mechanism in vision models decade ago. Oliva et al.~\cite{oliva2003top} propose a top-down attention model for object detection based on global scene configuration. Rutishauser et al.~\cite{rutishauser2004bottom} demonstrate that pure bottom-up attention can facilitate object recognition with information about the location, size and shape of objects. An integrated model of top-down and bottom-up attention is proposed in~\cite{navalpakkam2006integrated} to optimize target detection speed.
Such kind of attention mechanism has rarely been explored since the deep learning era. The most recent work is~\cite{girdhar2017attentional} , where top-down and bottom-up attention maps are combined as a classifier for action recognition. 

In this work, we devise a bidirectional attention module (BAM) with both top-down and bottom-up attention flows. The former is completely driven by the action recognition objective, while the latter is explicitly supervised by predicted saliency maps. Compared with~\cite{girdhar2017attentional}, BAM scales better in few-shot tasks since it works independently as an attention module instead of a classifier.

\noindent\textbf{Contrastive learning.} The idea of contrastive learning is first proposed for unsupervised representation learning~\cite{oord2018representation}. The essence is to pull similar samples closer and push dissimilar ones apart in the representation space with a contrastive loss~\cite{gutmann2010noise}. Recently proposed methods like MoCo~\cite{he2020momentum} and SimCLR~\cite{chen2020simple} have shown that unsupervised contrastive learning models are able to surpass supervised counterparts in image classification. Khosla et al.~\cite{khosla2020supervised} further extend contrastive learning to supervised scenarios and achieve further improvement. In this work, we adapt the contrastive loss to episodic meta-learning scenario and propose contrastive meta learning (CML), aiming to increase the discriminability of fine-grained action representations with massive contrastive power.

\vspace{-2mm}
\section{Method}

\noindent{\textbf{Problem definition.}} Following the standard definition of few-shot recognition problem~\cite{vinyals2016matching,snell2017prototypical,sung2018learning,zhu2018compound}, we separate the fine-grained action recognition dataset into a training set $D_{train}=\{(x_i, y_i)|y_i\in{C_{train}}\}$ and a test set $D_{test}=\{(x_i, y_i)|y_i\in{C_{test}}\}$, where $x_i$ is the i-th video with action label $y_i$. $C_{train}$ and $C_{test}$ are the disjoint training and test action categories, i.e. $C_{train}\cap{C_{test}}=\varnothing$. The definition of few-shot action recognition is training a model on $D_{train}$ to recognize actions in $D_{test}$ given only few support samples from $D_{test}$. Concretely, to perform a $N$-way $K$-shot $Q$-query task, a support set $\mathcal{S}$ and a query set $\mathcal{Q}$ are generated from $D_{test}$ as: 
\begin{equation}
    \mathcal{S}, \mathcal{Q}=\{(x_i, y_i)|y_i\in{C_{support}}\},
    \label{eqn:1}
\end{equation} 
where $C_{support}{\subseteq}C_{test}$ is the $N$ classes selected to be recognized. For each class in $C_{support}$, $K$ and $Q$ samples are randomly selected in the support and query sets respectively, resulting in $\lvert{\mathcal{S}}\rvert = N \times K$ and $\lvert{\mathcal{Q}}\rvert = N \times Q$. Models are then tested on $\mathcal{Q}$ only with the help from support samples in $\mathcal{S}$. 

In terms of model training, we adopt the episodic training strategy~\cite{snell2017prototypical,sung2018learning}, which trains the model episode by episode with a similar sampling method as the test task. In each training episode, we sample $\mathcal{S}$ and $\mathcal{Q}$ by randomly selecting $C_{support}$ from $C_{train}$ and optimize our model with the sampled data. 

\noindent{\textbf{Overall framework.}} As shown in Figure~\ref{fig:framework}, we first employ a backbone network to extract multi-scale spatio-temporal features from input videos. Specifically, we utilize a TSM~\cite{lin2019tsm} model with ResNet-50~\cite{he2016deep} architecture. For each input video clip, we extract spatio-temporal features at three different scales, which are respectively produced from \emph{conv3\_x}, \emph{conv4\_x} and \emph{conv5\_x} blocks of ResNet-50. Extracted features at each scale are then fed into a bidirectional attention module (BAM), where top-down (task-driven) and bottom-up (saliency-supervised) attention maps are summed and multiplied on input features. We utilize a model proposed in~\cite{li2019motion} to predict saliency volumes and down-sample them to multiple scales as saliency supervision. 
Next, we apply a three-level temporal pyramid network (TPN)~\cite{yang2020temporal} to aggregate the attended multi-scale features into a unified representation for each video. Finally, the proposed contrastive meta-learning (CML) is performed in this representation space for few-shot recognition. 

\begin{figure}[t]
  \centering
  \includegraphics[width=0.95\linewidth]{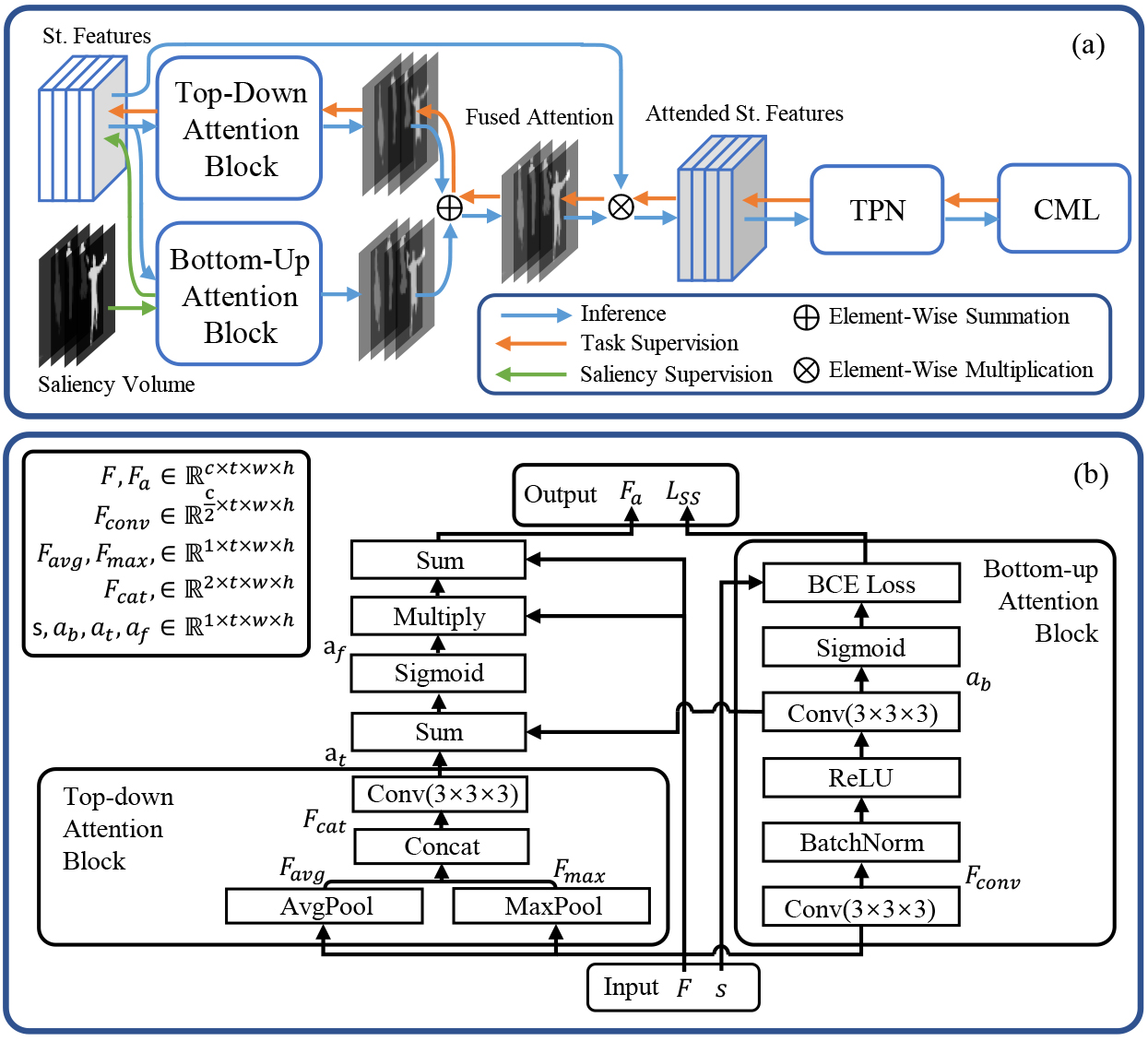}
  \vspace{-4mm}
  \caption{Illustration of the proposed bidirectional attention module (BAM). (a) Schema of information flows in BAM. Top-down and bottom-up attention blocks are supervised by task-driven and saliency-supervised signals, respectively. (b) Detailed structure of BAM.}
  \vspace{-6mm}
  \label{fig:dam}
\end{figure}

\subsection{Bidirectional Attention Module}
\label{subsec:bam}

Fine-grained action recognition requires differentiating between subtle motion details of fine-grained categories~\cite{shao2020finegym}. Unlike coarse-grained actions, which can be roughly distinguished by the background, the discriminative features of fine-grained actions are usually human/object-relevant, e.g. the motion trajectory of human body, the appearance of unique objects, etc. Human visual system has developed an effective attention mechanism to capture such subtle visual cues from large amounts of information~\cite{posner1990attention}. More specifically, 
human brain combines top-down task-driven signals with bottom-up salient stimuli together to facilitate complex visual tasks~\cite{ungerleider2000mechanisms,buschman2007top,baluch2011mechanisms}. To simulate this attention mechanism for our task, we exquisitely design a bidirectional attention module (BAM).  

For each scale of the spatio-temporal feature generated by the backbone, a BAM is employed to highlight fine-grained visual cues. As shown in Figure~\ref{fig:dam}(a), a BAM consists of both top-down and bottom-up attention blocks, which simulate attention signals from opposite directions. In the training phase, the top-down attention block is straightly supervised by the few-shot action recognition task, using gradients propagated from later procedures, i.e. TPN and CML. The bottom-up attention block is supervised by predicting saliency volumes from input features, conducting bottom-up salient stimuli. In the test phase, spatio-temporal attention maps generated from the two blocks are summed element-wisely to produce the fused attention map, which is then multiplied to the input to produce attended spatio-temporal features.

As illustrated in Figure~\ref{fig:dam}(b), we use $c$, $t$, $w$, $h$ to denote the size of feature maps in channel, temporal, width and height dimensions. Given input spatio-temporal feature map $F\in{\mathbb{R}^{c{\times}t{\times}w{\times}h}}$ and down-sampled saliency volume $s\in{\mathbb{R}^{1{\times}t{\times}w{\times}h}}$, BAM outputs attended feature map $F_{a}\in{\mathbb{R}^{c{\times}t{\times}w{\times}h}}$ and saliency supervision loss $L_{ss}$. We generate $F_a$ as:
\begin{equation}
    \begin{split}
    & F_a=F+F\otimes{\sigma(f_{t}(F)+f_{b}(F))}\\
    & \quad =F+F\otimes{\sigma(a_{t}+a_{b})}
    \end{split},
\end{equation}
where $f_{t}$ and $f_{b}$ are the functions of top-down and bottom-up attention blocks respectively. $a_{t},a_{b}\in{\mathbb{R}^{1{\times}t{\times}w{\times}h}}$ represent corresponding output attention maps. $\sigma$ is the sigmoid activation function. $\otimes$ denotes element-wise multiplication. We apply a residual attention~\cite{wang2017residual} design to stabilize training gradients, i.e. $F$ is added to the attended version of itself with a residual path. 

To efficiently generate the top-down attention, we adopt a similar attention block as CBAM~\cite{woo2018cbam} for $f_t$. Average- and max-pooling operations are applied on $F$ channel-wisely. The outputs are concatenated to form an efficient feature descriptor $F_{cat}\in{\mathbb{R}^{2{\times}t{\times}w{\times}h}}$, which is convolved by a 3D convolution layer with the kernel size of $3\times3\times3$. For bottom-up attention block $f_b$, in order to predict $s$ from $F$, we devise a lightweight decoder containing two $3\times3\times3$ convolution layers, with the first one followed by a batch normalization layer and a ReLU activation. We then detach the task-related gradients and impose the saliency supervision loss $L_{ss}$ on $a_b$ as:
\begin{equation}
    L_{ss}=s{\cdot}log({\sigma}(a_{b}))+(1-s){\cdot}log(1-{\sigma}(a_{b})),
\end{equation}
which takes the form of binary cross-entropy (BCE) loss. $L_{ss}$ enforces $a_{b}$ to reconstruct $s$ by emphasizing the salient areas and suppressing the others. 

In each training episode, $L_{ss}$ is imposed on all samples from both support and query sets, which can be regarded as introducing an additional saliency prediction task into our framework. Notably, the saliency input is only needed in the training phase. In the test phase, the predicted saliency map (i.e. $\sigma(a_b)$) can be utilized as a by-product to indicate salient humans and objects in videos as shown in Section~\ref{subsec:results}.

\subsection{Contrastive Meta-Learning}
\label{subsec:cml}

\begin{figure}[t]
  \centering
  \includegraphics[width=0.9\linewidth]{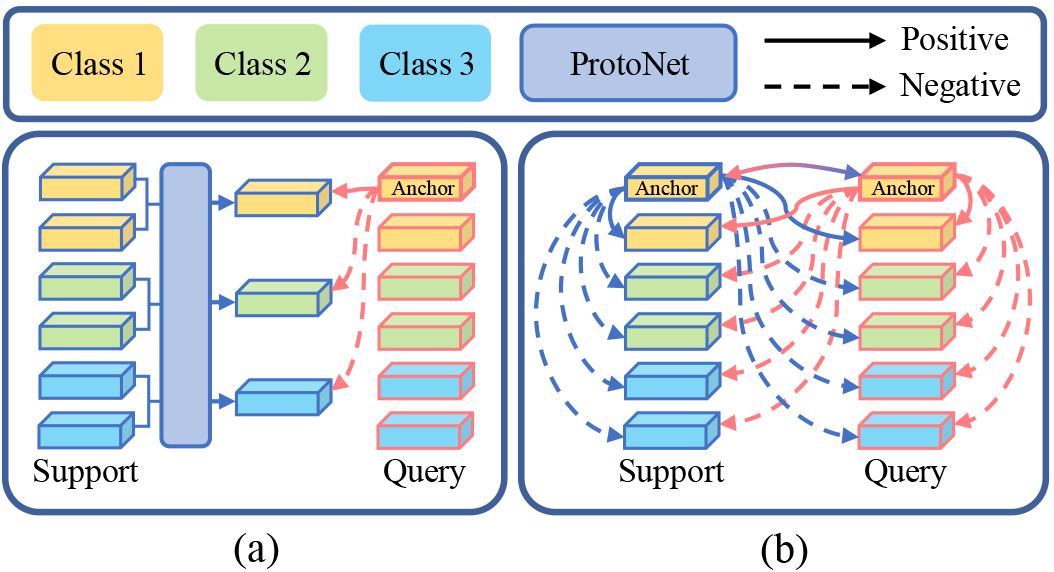}
  \vspace{-4mm}
  \caption{Comparison of (a) ProtoNet-based meta-learning and (b) proposed contrastive meta-learning (CML) on a 3-way 2-shot 2-query task. Given the same number of samples in each training episode, CML creates far more contrastive pairs than the ProtoNet-based method.}
  \vspace{-5mm}
  \label{fig:cml}
\end{figure}

As shown in Figure~\ref{fig:cml}(a), most state-of-the-art few-shot action recognition methods~\cite{fu2020depth,zhu2021few,perrett2021temporal} utilize the prototypical network (ProtoNet)~\cite{snell2017prototypical} for meta-learning. In each training episode, representations of support samples are utilized to build a classifier.  The model is then optimized by classifying query samples with a cross-entropy (CE) loss. For multi-shot ($N>1$) recognition problems, ProtoNet is used to fuse multiple support representations for each class. Given the support set $\mathcal{S}$ and query set $\mathcal{Q}$, ProtoNet-based models aim to optimize to the classification loss:
\begin{equation}
    L_{ce} = -\sum_{x_i\in{\mathcal{Q}}}{log\frac{exp(\lVert{R_i,P_{l(i)}}\rVert/\tau)}{\sum_{j=1}^{N}{exp(\lVert{R_i,P_j}\rVert/\tau})}},
    \label{eqn:2}
\end{equation}
where $R_i=f_{\theta}(x_i)$ is the representation of video $x_i$ produced by a model $f(\cdot)$ with parameters $\theta$. $P_j$ is the prototype representation of the j-th class in $N$ classes. The few-shot label of query $x_i$ is given by $l(i)$. We use $\lVert{\cdot}\rVert$ to denote the distance similarity function, which is used to compute the similarity between representations. $\tau$ is the scalar temperature parameter~\cite{he2020momentum}. Following the definition of ProtoNet, $P_j$ is calculated by averaging all support video representations in class $j$ as:
\begin{equation}
    P_j=\frac{\sum_{{x_i}\in{\mathcal{S}_j}}{R_i}}{\lvert{\mathcal{S}_j}\rvert},
    \label{eqn:3}
\end{equation}
where $\mathcal{S}_j=\{(x_{i}, y_{i})|x_{i}\in{\mathcal{S}},y_{i}=j\}$ contains all support samples in class $j$.

Although widely adopted in few-shot action recognition, this ProtoNet-based method is not optimal for fine-grained actions. As shown in Figure~\ref{fig:intro}, the inter-class variance of fine-grained actions is relatively low compared with coarse-grained actions. Therefore, samples from different categories are very likely to be tangled with each other in the representation space, making it tricky to distinguish query samples simply by the averaged prototypes. 
To generate more discriminative representations, ProtoNet requires a larger episodic training batch~\cite{snell2017prototypical,sung2018learning}, e.g. a 10-way training batch for a 5-way test task, which is prohibitively expansive for video data. It is thus very desirable to make the best of limited training samples in each episode without expanding the batch size.

Recent studies~\cite{he2020momentum,chen2020simple,khosla2020supervised} have shown that contrastive learning is adept at generating discriminative representations under limited supervision. We hence propose contrastive meta-learning (CML), which employs a contrastive loss to learn video representations under the episodic meta-learning setting.
As shown in Figure~\ref{fig:cml}(b), CML greatly improves the discriminability of video representations by making full use of the potential contrastive pairs. Given any training sample from $\mathcal{S}$ and $\mathcal{Q}$, it is contrasted to all the other samples by taking those of the same class as positives and others as negatives. The loss function of CML is defined as:
\begin{small}
\begin{equation}
    L_{cml}=\sum_{x_i\in{\mathcal{A}}}{\frac{-1}{\lvert{\mathcal{P}(i)}\rvert}}\sum_{x_j\in{\mathcal{P}(i)}}{log\frac{exp(\lVert{R_i},R_j\rVert/\tau)}{\sum\limits_{{x_k}\in{\mathcal{A}^-}}{exp(\lVert{R_i},R_k\rVert/\tau)}}},
    \label{eqn4}
\end{equation}
\end{small}
where $\mathcal{A}=\mathcal{S}\cup{\mathcal{Q}}$ contains all samples from support and query sets and $\mathcal{A}^-=\mathcal{A}\setminus{x_i}$. The positive set for anchor $x_i$ is denoted as $\mathcal{P}(i)=\{(x_j, y_j)|x_j\in{\mathcal{A}^-},y_{j}=y_i\}$. Accordingly, the negative set is $\mathcal{A}^-{\setminus{\mathcal{P}(i)}}$. 

In the test phase, CML straightly compares query representations with every support representation in each class to make the few-shot prediction, which is formulated as:
\begin{equation}
   Pred(x_i) = \mathop {argmax}_{c \in N} \sum_{{x_j} \in {\mathcal {S}},l(j) = c} {\lVert{R_i},{R_j}\rVert},
    \label{eqn5}
\end{equation}
where the predicted class $c$ maximizes the overall similarity between $R_i$ and every $R_j$ in class $c$. Since we increase discriminability of video representations with contrastive learning and remove the averaged prototype, CML is more robust to low inter-class variance of fine-grained action categories. Compared with the ProtoNet-based method, CML unleashes much more contrastive power from limited samples in each training episode. Here we mathematically demonstrate this advantage of CML.

For a $N$-way $K$-shot $Q$-query meta-training episode, we denote the number of positives and negatives as $n_{pos}$ and $n_{neg}$. The number of total contrastive pairs is thus $n_{all}=n_{pos}+n_{neg}$. Taking $L_{ce}$ as a special case of contrastive loss~\cite{gutmann2010noise}, we have:
\begin{equation}
    n_{pos}=\lvert{\mathcal{Q}}\rvert,\ {n_{neg}}=\lvert{\mathcal{Q}}\rvert(N-1),
\end{equation}
i.e. 
\begin{equation}
    n_{pos}=NQ,\ {n_{neg}}=NQ(N-1), \ n_{all}=N^2Q.
\end{equation}
In comparison, the numbers of contrastive pairs for $L_{cml}$ are:
\begin{equation}
    n_{pos}=\lvert{\mathcal{A}}\rvert{\lvert{\mathcal{P}(i)}\rvert},\ {n_{neg}}=\lvert{\mathcal{A}}\rvert{(N-1)(K+Q)},
\end{equation}
i.e.
\begin{equation}
\begin{split}
    & n_{pos}=N(K+Q)(K+Q-1),\\ 
    & {n_{neg}}=N(N-1)(K+Q)^2,\\
    & n_{all} = N^2(K+Q)^2-N(K+Q).
\end{split}
\end{equation}

Let $n_{all}(CE)$ and $n_{all}(CML)$ denote the number of total contrastive pairs for $L_{ce}$ and $L_{cml}$ respectively, the former one is a three degree polynomial w.r.t. $N$ and $Q$ while the latter one is a four degree polynomial w.r.t. $N$, $K$ and $Q$. Given that $N,K,Q\in{\mathbb{N}^+}$, we can easily deduce $n_{all}{(CML)} \gg n_{all}{(CE)}$. This indicates that with the same number of samples in each training episode, CML creates far more contrastive pairs than the ProtoNet-based method, thus greatly increasing the contrastive power of our model.
Since CML is performed in the representation space, the increment of computational cost from $L_{cml}$ is marginal compared with the cost of representation generation. In fact, by removing the ProtoNet module, CML runs faster in both training and test phases. More details are given in our supplementary material.

\subsection{Overall Objective}

With the saliency supervision loss $L_{ss}$ and the contrastive meta-learning loss $L_{cml}$ defined above, the overall loss function of our framework is formulated as:
\begin{equation}
    L ={\alpha}\sum\nolimits_{i=1}^3 L_{ss(i)}+{\beta}L_{cml},
\end{equation}
where $L_{ss(i)}$ is the saliency supervision loss of the i-th spatio-temporal scale. $\alpha$ and $\beta$ are the corresponding loss weights which are empirically set to 0.1 and 1. Our model is trained in an end-to-end manner. In each training episode, $L_{ss}$ and $L_{cml}$ are imposed simultaneously on samples from both support and query sets.

\vspace{-1.5mm}
\section{Experiments}
\label{sec:exp}

\subsection{Benchmark Protocols}

We evaluate our model and the competitor methods on two recently proposed large-scale fine-grained action recognition datasets, namely FineGym~\cite{shao2020finegym} and HAA500~\cite{chung2020haa500}. All models are tested on 5-way 1/3/5-shot recognition tasks. Since the training and test protocols are quite influential on the result of FSL problems~\cite{snell2017prototypical,zhang2020few}, we establish specific benchmark protocols~\footnote{\url{https://github.com/acewjh/FSFG}} upon the two datasets, including the split of training/test action categories and the number of samples in each training/test episode. 

\textbf{FineGym} contains hierarchical annotations of fine-grained actions in gymnastic events. There are two levels of fine-grained categories, i.e. Gym99 and Gym288, consisting of more than 34k (99 classes) and 38k (288 classes) samples respectively. For Gym99, we take a 79/20 split for the training/test categories. For Gym288, the traning/test categories are set to 231/57 for 1-shot recognition task. Since some actions in Gym288 do not have sufficient support samples for multi-shot tasks, we get rid of them, resulting in 218/54 and 197/50 splits for 3-shot and 5-shot tasks.

\textbf{HAA500} is comprised of 10k human-centric action videos from 500 fine-grained classes. The actions are from different areas, including sports, instrument performance and daily actions. We set the training/test class split on HAA500 to 400/100.

To fairly compare different models, in the training phase, we fix the number of query samples ($Q$) to 5 in each episode across 1/3/5-shot tasks. The number of way ($N$) and shot ($K$) are consistent with corresponding test tasks. In the test phase, $Q$ is also set to 5 in each episode. Every evaluation is made by randomly sampling 6,000 test episodes. We take the mean accuracy over all test episodes along with the $95\%$ confidence interval as evaluation metrics.

\vspace{-1.5mm}
\subsection{Implementation Details}

We implement our model with PyTorch. The backbone network is initialized with ImageNet pretrained weights. The TPN module is trained from scratch with random initialization. We utilize the officially released model in~\cite{li2019motion} to predict saliency maps. Following the sampling strategy of TSM~\cite{lin2019tsm}, we sample 8 frames from each video as the input clip, with each frame resized and cropped to $224\times224$. We adopt the cosine similarity function~\cite{he2020momentum} in $L_{cml}$. Our model is trained on two Nvidia RTX 3090 GPUs. We utilize the SGD optimizer with a initial learning rate of 0.005, which decays at the 75 and 125 epochs by 0.1. The dropout rate is set to 0.5. The total training episodes are 40,000 and 50,000 on FineGym and HAA500. 

\vspace{-4mm}
\subsection{Competitor Methods} 

Challenging competitor methods are selected in comparison with our approach from three different measures:

\noindent\textbf{State-of-the-art action recognition methods.} 
SlowFast~\cite{feichtenhofer2019slowfast} represents the state-of-the-art model for general action recognition. We select the R-50 backbone and adapt the model to FSL with ProtoNet and CML respectively, denoted as \emph{SlowFast} and \emph{SlowFast++}. TRX~\cite{perrett2021temporal} is the state-of-the-art few-shot action recognition model, which takes frame tuples from a cardinality set ($\Omega$) as input. We evaluate TRX with $\Omega=\{2,3\}$ following the best practice.

\noindent\textbf{Attention modules for action recognition.} Two attention modules designed for fine-grained action recognition, i.e. attentional pooling~\cite{girdhar2017attentional} and RRA~\cite{zhu2018fine}, are compared with our BAM. Since the attentional pooling module works as a classifier, we directly employ it on backbone features with class-specific and class-agnostic attention maps generated from support and query samples respectively. RRA is evaluated as a substitute of BAM with the same framework configurations. We refer to RRA combined with ProtoNet and CML as \emph{RRA} and \emph{RRA++}, respectively.

\noindent\textbf{Meta-learning methods.} In comparison with the proposed CML, three classic meta-learning methods are evaluated, namely MatchingNet~\cite{vinyals2016matching}, ProtoNet~\cite{snell2017prototypical} and RelationNet~\cite{sung2018learning}. All three methods are implemented on the same backbone and TPN architectures as ours. MatchingNet utilizes a bidirectional LSTM to generate fully conditional embeddings (FCE). The relation module in RelationNet is built following configurations from the original paper.

\renewcommand\arraystretch{0.92}
\begin{table*}[t]
\small
\caption{Performance comparison on benchmarks of Gym99, Gym288 and HAA500. Results are reported by the averaged accuracy (\%) over 6,000 randomly sampled test episodes with 95\% confidence intervals.} 
\vspace{-4mm}
\label{tab:results}
\begin{center}
\setlength{\tabcolsep}{1.4mm}{
\begin{tabu}{l|ccc|ccc|ccc}
\tabucline[1.25pt]{-}
\multicolumn{1}{l|}{\multirow{2}{*}{Method}} & \multicolumn{3}{c|}{Gym99} & \multicolumn{3}{c|}{Gym288} & \multicolumn{3}{c}{HAA500} \\
\multicolumn{1}{c|}{}                        & 1-shot  & 3-shot  & 5-shot & 1-shot  & 3-shot  & 5-shot  & 1-shot  & 3-shot  & 5-shot  \\ \hline
SlowFast~\cite{feichtenhofer2019slowfast} & $87.44\pm0.33$ & $89.27\pm0.30$ & $90.15\pm0.28$ & $81.54\pm0.45$ & $83.11\pm0.42$ & $82.69\pm0.43$ & $82.15\pm0.36$ & $83.31\pm0.34$ & $83.85\pm0.34$\\
SlowFast++~\cite{feichtenhofer2019slowfast} & $90.51\pm0.28$ & $91.13\pm0.26$ & $91.94\pm0.23$ & $83.43\pm0.38$ & $85.79\pm0.34$ & $86.24\pm0.32$ & $83.07\pm0.33$ & $86.26\pm0.30$ & $87.41\pm0.28$\\
TRX~\cite{perrett2021temporal} & $79.68\pm0.49$ & $84.15\pm0.47$ & $86.58\pm0.45$ & $76.62\pm0.53$ & $77.39\pm0.54$ & $77.98\pm0.51$ & $73.24\pm0.64$ & $76.84\pm0.58$ & $76.81\pm0.60$\\ \hline
Attn. Pooling~\cite{girdhar2017attentional} & $75.37\pm0.61$ & $76.39\pm0.62$ & $78.24\pm0.61$ & $71.52\pm0.72$ & $71.87\pm0.68$ & $73.12\pm0.67$ & $68.18\pm0.75$ & $70.29\pm0.72$ & $71.21\pm0.73$\\
RRA~\cite{zhu2018fine} & $87.08\pm0.31$ & $90.14\pm0.27$ & $92.69\pm0.24$ & $80.85\pm0.37$ & $83.45\pm0.38$ & $85.84\pm0.36$ & $70.22\pm0.35$ & $78.57\pm0.29$ & $83.35\pm0.25$ \\
RRA++~\cite{zhu2018fine} & $93.07\pm0.25$ & $93.80\pm0.22$ & $94.19\pm0.22$ & $87.78\pm0.34$ & $88.80\pm0.32$ & $88.79\pm0.34$ & $84.39\pm0.29$ & $89.82\pm0.21$ & $90.78\pm0.20$\\
BAM (ours) & $89.38\pm0.27$ & $90.21\pm0.26$ & $93.10\pm0.24$ & $85.67\pm0.35$ & $89.35\pm0.33$ & $88.81\pm0.35$ & $82.96\pm0.36$ & $87.56\pm0.33$ & $89.04\pm0.28$\\
\hline
MatchingNet~\cite{vinyals2016matching} & $78.27\pm0.54$ & $79.42\pm0.53$ & $81.33\pm0.50$ & $72.07\pm0.68$ & $74.51\pm0.61$ & $74.78\pm0.62$ & $70.27\pm0.71$ & $73.29\pm0.67$ & $74.13\pm0.66$ \\
ProtoNet~\cite{snell2017prototypical} & $88.37\pm0.30$ & $90.23\pm0.28$ & $90.83\pm0.26$ & $83.69\pm0.40$ & $84.49\pm0.37$ & $84.23\pm0.38$ & $82.41\pm0.34$ & $85.25\pm0.33$ & $84.59\pm0.35$ \\ 
RelationNet~\cite{sung2018learning} & $89.32\pm0.31$ & $90.18\pm0.30$ & $91.25\pm0.28$ & $84.65\pm0.43$ & $85.44\pm0.40$ & $85.41\pm0.38$ & $82.39\pm0.44$ & $85.68\pm0.42$ & $85.21\pm0.41$\\ 
CML (ours) & $91.98\pm0.26$ & $90.43\pm0.27$ & $94.13\pm0.22$ & $86.11\pm0.35$ & $88.34\pm0.34$ & $89.08\pm0.33$ & $84.10\pm0.28$ & $89.55\pm0.22$ & $90.63\pm0.19$        \\
\hline
BAM+CML (ours) & $\mathbf{93.46\pm0.24}$ & $\mathbf{93.87\pm0.22}$ & $\mathbf{94.45\pm0.21}$ & $\mathbf{89.26\pm0.31}$ & $\mathbf{90.16\pm0.30}$ & $\mathbf{90.97\pm0.30}$ & $\mathbf{85.45\pm0.27}$ & $\mathbf{90.68\pm0.21}$ & $\mathbf{91.84\pm0.19}$ \\
\tabucline[1.25pt]{-}
\end{tabu}}
\vspace{-4mm}
\end{center}
\end{table*}

\vspace{-4mm}
\subsection{Results}
\label{subsec:results}

\noindent\textbf{Benchmark observations.} Table~\ref{tab:results} presents the complete benchmark results. Our method consistently outperforms other competitors across all few-shot tasks on Gym99, Gym288 and HAA500. The major observations can be concluded as below: 

\noindent\textbf{(1)} 
Our method significantly exceeds SlowFast and TRX, the advantage is especially obvious on Gym288 and HAA500, demonstrating that models designed for coarse-grained action recognition are inappropriate for fine-grained tasks. Since TRX only takes sparsely sampled frame tuples as input, it is very likely to miss fine-grained action details, which explains its inferior performance. 

\noindent\textbf{(2)} 
Although~\cite{girdhar2017attentional} also adopts a top-down and bottom-up attention design, it does not work well on few-shot recognition tasks. Because it utilizes class-specific attention maps as classifiers, which are tricky to learn with few support samples in each training episode. RRA shows better performance, while still largely outperformed by our BAM.

\noindent\textbf{(3)} MatchingNet shows the worst performance in all meta-learning methods. Though ProtoNet and RelationNet achieve further improvement, our CML outperforms both methods by a large margin. Notably, both methods failed to increase the accuracy from 3-shot to 5-shot on Gym288 and HAA500. This indicates that the discriminability of prototype representations is hard to improve with more support samples. In comparison, CML consistently improves the accuracy from 1-shot to 5-shot tasks.

\noindent\textbf{(4)}
CML also shows better performance combined with other architectures (\emph{SlowFast++}, \emph{RRA++}) in comparison with ProtoNet, demonstrating its robustness across different model designs.

\begin{figure}[t]
  \centering
  \includegraphics[width=0.95\linewidth]{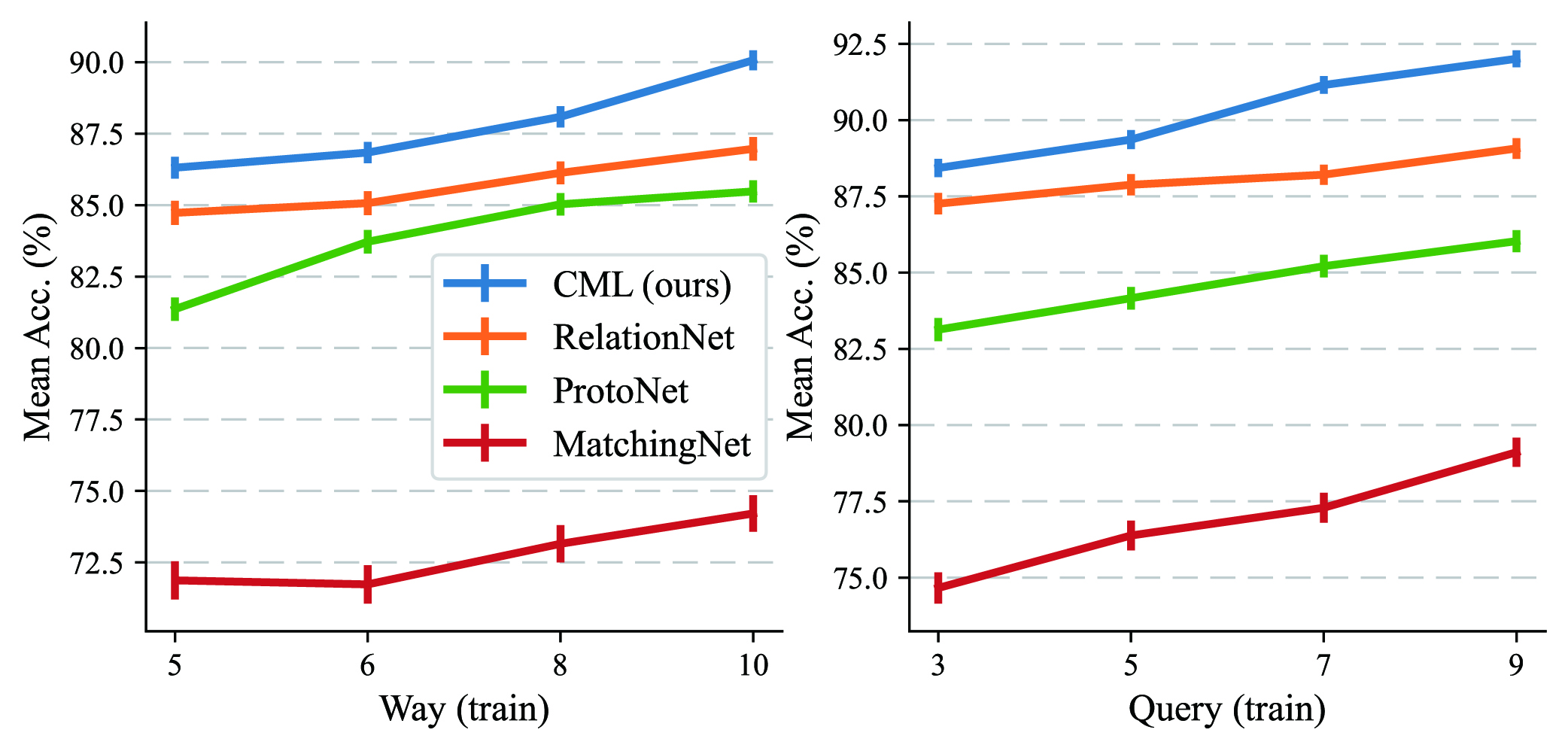}
  \vspace{-3mm}
  \caption{Left: Accuracy on 5-way, 2-shot, 2-query tasks of models trained with 5/6/8/10-way, 2-shot, 2-query episodes. Right: Accuracy on 3-way, 2-shot, 2-query tasks of models trained with 3-way, 2-shot, 3/5/7/9-query episodes.}
  \vspace{-3mm}
  \label{fig:ml}
\end{figure}

\begin{figure}[t]
  \centering
  \includegraphics[width=1\linewidth]{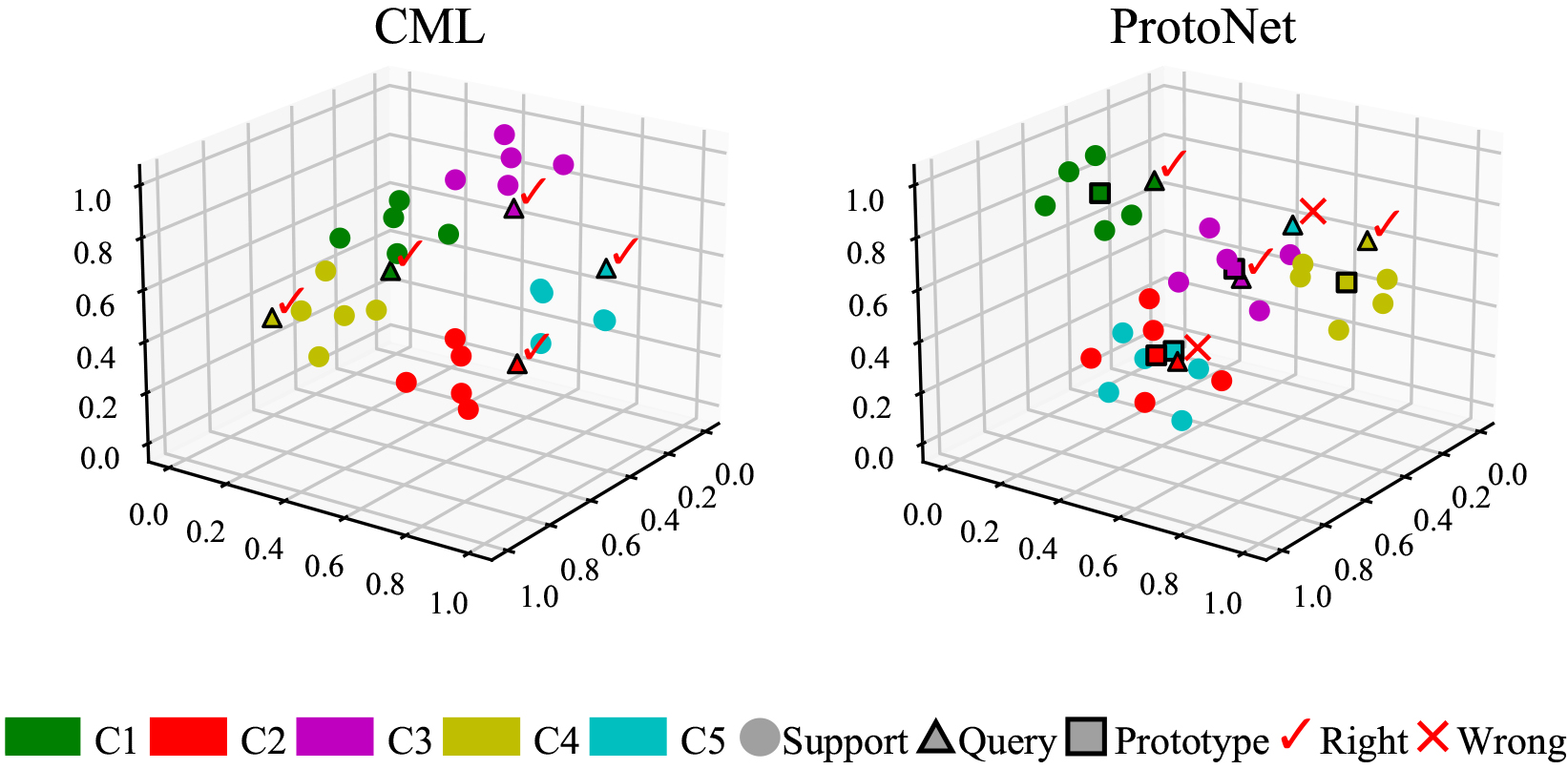}
  \vspace{-7mm}
  \caption{3D projections of video representations generated by CML and ProtoNet in a same 5-way 5-shot 1-query task.} 
  \vspace{-5mm}
  \label{fig:rep}
\end{figure}

\begin{figure*}[t]
\begin{center}
   \includegraphics[width=0.95\textwidth]{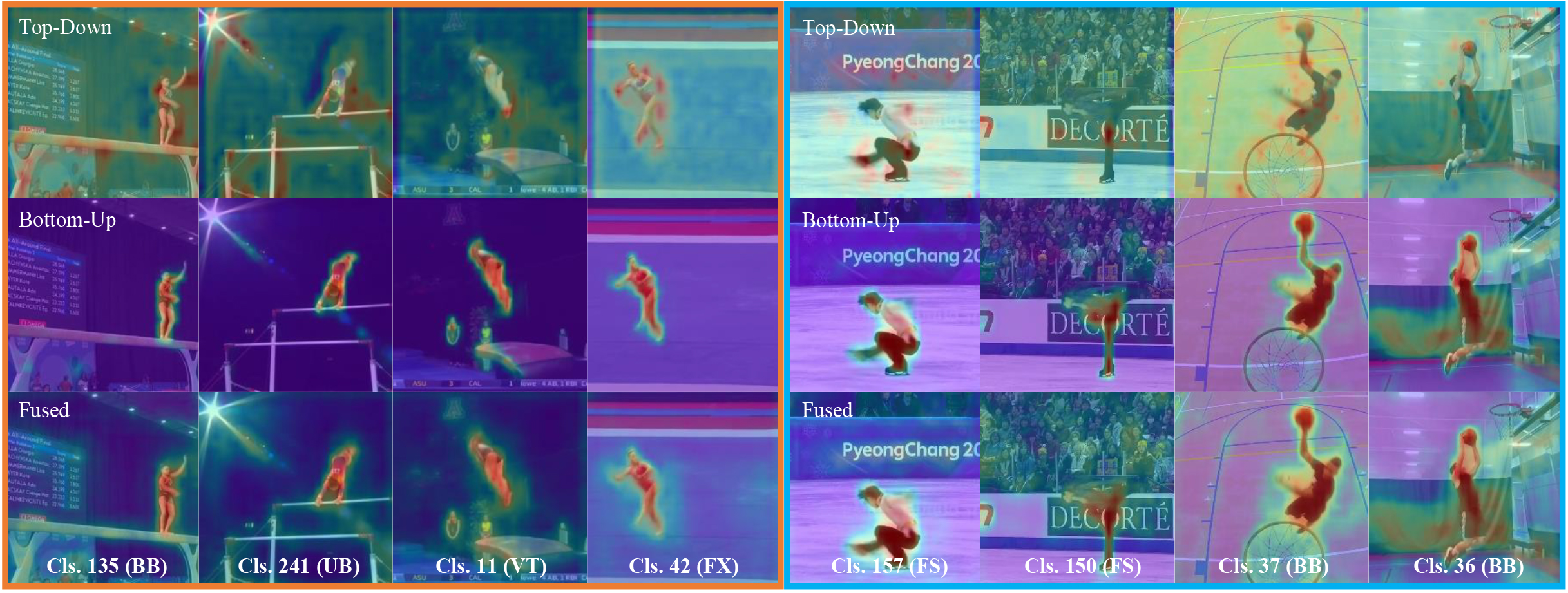}
\end{center}
   \vspace{-4mm}
   \caption{Visualization of attention maps generated by BAM. Four samples from each of Gym288 (left) and HAA500 (right) datasets are shown with top-down, bottom-up and fused attention maps in different rows. Specific action labels are given at the bottom. Top-down attention focuses on class-related regions including critical body parts, sports equipments and discriminative backgrounds. Bottom-up attention captures salient visual cues like athletes and objects in motion. 
   }
   \vspace{-3mm}
\label{fig:attention}
\end{figure*}

\noindent\textbf{Attention visualization.} 
We visualize samples of BAM generated attention maps from Gym288 and HAA500 in Figure~\ref{fig:attention}. For Gym288, the samples are from four basic gymnastic events, i.e. balance beam (BB), uneven bars (UB), vault (VT) and floor exercise (FX). For HAA500, fine-grained actions from figure skating (FS) and basketball (BB) are shown. 

Supervised by the action recognition task, top-down attention learns to focus on class-related information like sports equipments and discriminative backgrounds, e.g. balance beams, uneven bars, ice rinks and basketball hoops. Moreover, critical body parts of athletes are also captured by top-down attention. For instance, to perform \emph{sit spin} (cls. 157) in figure skating, the skater squats down and spins with one foot in the air. The attention map hence focuses on feet and hips. To perform \emph{donut spin} (cls. 150), the skater spins on a single leg, which is attended accordingly. 

In comparison, bottom-up attention concentrates on salient visual cues like athletes and objects (e.g. basketball) in motion, since it is supervised by saliency volumes. Class-agnostic salient areas are highlighted with clear boundaries. As a by-product, the resulting attention maps can be utilized to facilitate other video tasks like video matting and retargeting.

Combining two attention maps together, fused attention effectively captures the informative spatio-temporal regions containing subtle action details, with maximal attention weights assigned to the major performer and high attention weights to discriminative surroundings. Uninformative regions are suppressed otherwise.

\noindent\textbf{Comparison of meta-learning methods.} To evaluate the capacity of different meta-learning methods on fine-grained actions, we conduct two comparative experiments on Gym288. The evaluated models only differ in meta-learning methods. In the first experiment, we explore the influence of way numbers ($N$) in training episodes. Models are tested on 5-way, 2-shot, 2-query tasks and trained with 5/6/8/10-way, 2-shot, 2-query batches. In the second experiment, the impact of query numbers ($Q$) is investigated. Models are tested on 3-way, 2-shot, 2-query tasks and trained with 3-way, 2-shot, 3/5/7/9-query batches. The results are illustrated in Figure~\ref{fig:ml}.

It can be observed that for both experiments, CML shows higher growth rate of accuracy compared with RelationNet and ProtoNet. It demonstrates that CML has higher learning capacity on fine-grained data,
consisting with the theoretical analysis in Section~\ref{subsec:cml}. Although the accuracy of MatchingNet also increases rapidly, the margin between CML and MatchingNet is significant. Moreover, in the first experiment, the slope of CML increases with more training classes, whereas the one of ProtoNet decreases. This implies that CML consistently improves feature discriminability with increasing training classes, while the marginal effect decreases for ProtoNet.

We visualize 3D projections of video representations generated by CML and ProtoNet in a same 5-way 5-shot 1-query task in Figure~\ref{fig:rep} with t-SNE~\cite{van2008visualizing} algorithm. 
All query samples are correctly classified by CML, while two are miss-classified by ProtoNet. The mistakes of ProtoNet are caused by tangled prototypes (C2) and deviated representations (C5), showing the weakness of ProtoNet on low inter-class variance data. In comparison, CML generates video representations with clear inter-class separations and compact intra-class distributions, even for very similar categories like C2 and C5.

\begin{figure}[t]
  \centering
  \includegraphics[width=0.9\linewidth]{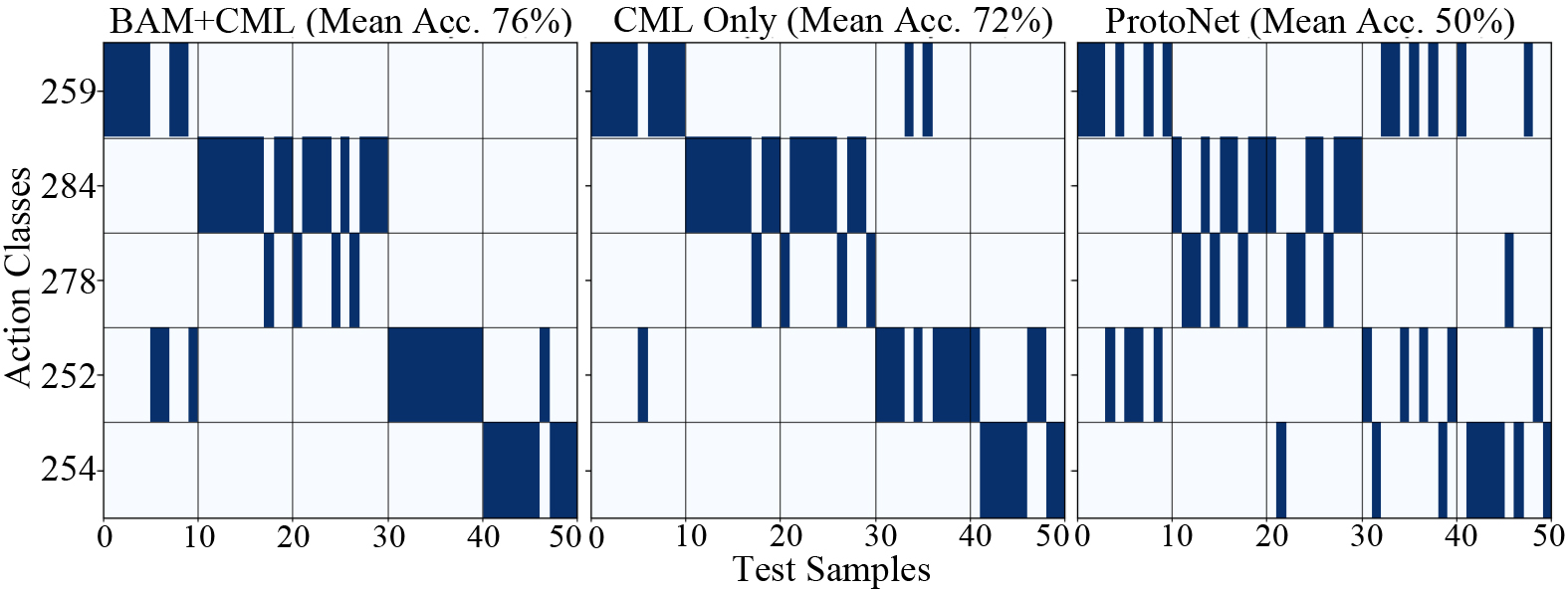}
  \vspace{-4mm}
  \caption{Confusion matrices of ablative results on a hard fine-grained task. Mean accuracies are shown at the top.}
  \vspace{-6mm}
  \label{fig:cm}
\end{figure}

\subsection{Ablation Study}
\label{ablation}
\noindent\textbf{Effectiveness of multi-scale features.} To validate the effectiveness of multi-scale spatio-temporal features, we ablate backbone features at the first (\emph{conv3\_x}) and second (\emph{conv4\_x}) scales, which are referred to as \emph{Scale\{x\}}, with \emph{x} denoting the remaining scales. For \emph{Scale\{2,3\}} model, a two-level TPN is utilized. For \emph{Scale\{3\}} model, the single scale feature map is directly employed without TPN. As Table~\ref{tab:ablation} shows, the ablation of the first feature scale leads to large performance drop ($2.20\%$, $3.65\%$, $2.09\%$) on all three benchmarks. Since features at the first scale has the highest spatio-temporal resolution, thus benefits the most from attention refinement.

\noindent\textbf{Efficacy of bidirectional attention.} We also conduct ablative experiments to assess the efficacy of bidirectional attention. The top-down and bottom-up attention blocks are respectively ablated from BAM. Table~\ref{tab:ablation} shows that bottom-up attention creates more enhancement than top-down attention across three benchmarks ($0.58\%$, $1.83\%$, $0.98\%$ vs. $-0.17\%$, $0.66\%$, $0.72\%$), because it introduces additional saliency supervision. Although top-down attention seems to fail on Gym99, it contributes on Gym288 and HAA500. In summary, the contributions of both attention blocks are more obvious with more fine-grained categories, implying that actions at finer granularities need more attention to be distinguished.

\noindent\textbf{Ablation results on hard fine-grained tasks.} We further evaluate the performance of BAM and CML on hard fine-grained tasks. A 5-way 5-shot 10-query test task is sampled within the UB event of Gym288. The selected actions are thus extremely similar mutually. We successively ablate BAM and CML from our framework and present the confusion matrices of test results in Figure~\ref{fig:cm}. Without BAM and CML, the hard fine-grained samples are severely confused, e.g. \emph{clear pike circle backward with 1 turn to handstand} (cls. 252) and \emph{stalder forward with 0.5 turn to handstand} (cls. 259). This result demonstrates that BAM and CML greatly improve discriminability of hard fine-grained actions.

\renewcommand\arraystretch{0.8}
\begin{table}[t]
\small
\caption{Results of ablative experiments. Models are uniformly tested on 5-way 1-shot 5-query tasks.}
\vspace{-3mm}
\label{tab:ablation}
\begin{center}
\begin{tabu}{l|c|c|c}
\tabucline[1.25pt]{-}
Method         & Gym99                 & Gym288                & HAA500               \\ \hline
BAM+CML (ours) & $94.45\pm0.21$ & $\mathbf{90.97\pm0.30}$ & $\mathbf{91.84\pm0.19}$ \\ \hline
Scale\{2,3\} & $92.25\pm0.28$ & $87.32\pm0.35$ & $89.75\pm0.26$ \\
Scale\{3\} & $91.43\pm0.28$ & $86.19\pm0.38$ & $89.06\pm0.27$ \\ \hline
w/o Top-Down & $\mathbf{94.62\pm0.22}$ & $90.31\pm0.30$ & $91.12\pm0.21$ \\
w/o Bottom-Up & $93.87\pm0.24$ & $89.14\pm0.32$ & $90.86\pm0.20$ \\
\tabucline[1.25pt]{-}
\end{tabu}
\vspace{-5mm}
\end{center}
\end{table}

\section{Conclusion}

We propose the few-shot fine-grained action recognition problem in this work, which is spawned from real-world demands. To effectively capture subtle details of fine-grained actions, we devise a human vision inspired bidirectional attention module (BAM), where task-driven and saliency-supervised signals are combined to highlight informative spatio-temporal regions. To generate discriminative representations for low inter-class variance data, we propose contrastive meta-learning (CML), making full use of contrastive pairs in each training episode. Specific benchmark protocols are established to fairly compare different models. Extensive experiments demonstrate our superiority and validate the efficacy of different components. In the future, we intend to extend our framework to other fine-grained action analysis tasks like action assessment.

\begin{acks}
This work was supported by National Natural Science Foundation of China (Grant No. U20B2069).
\end{acks}


\bibliographystyle{ACM-Reference-Format}
\bibliography{ref}


\begin{thebibliography}{53}


\ifx \showCODEN    \undefined \def \showCODEN     #1{\unskip}     \fi
\ifx \showDOI      \undefined \def \showDOI       #1{#1}\fi
\ifx \showISBNx    \undefined \def \showISBNx     #1{\unskip}     \fi
\ifx \showISBNxiii \undefined \def \showISBNxiii  #1{\unskip}     \fi
\ifx \showISSN     \undefined \def \showISSN      #1{\unskip}     \fi
\ifx \showLCCN     \undefined \def \showLCCN      #1{\unskip}     \fi
\ifx \shownote     \undefined \def \shownote      #1{#1}          \fi
\ifx \showarticletitle \undefined \def \showarticletitle #1{#1}   \fi
\ifx \showURL      \undefined \def \showURL       {\relax}        \fi
\providecommand\bibfield[2]{#2}
\providecommand\bibinfo[2]{#2}
\providecommand\natexlab[1]{#1}
\providecommand\showeprint[2][]{arXiv:#2}

\bibitem[\protect\citeauthoryear{Baluch and Itti}{Baluch and Itti}{2011}]%
        {baluch2011mechanisms}
\bibfield{author}{\bibinfo{person}{Farhan Baluch} {and}
  \bibinfo{person}{Laurent Itti}.} \bibinfo{year}{2011}\natexlab{}.
\newblock \showarticletitle{Mechanisms of top-down attention}.
\newblock \bibinfo{journal}{\emph{Trends in Neurosciences}}
  \bibinfo{volume}{34}, \bibinfo{number}{4} (\bibinfo{year}{2011}),
  \bibinfo{pages}{210--224}.
\newblock


\bibitem[\protect\citeauthoryear{Buschman and Miller}{Buschman and
  Miller}{2007}]%
        {buschman2007top}
\bibfield{author}{\bibinfo{person}{Timothy~J Buschman} {and}
  \bibinfo{person}{Earl~K Miller}.} \bibinfo{year}{2007}\natexlab{}.
\newblock \showarticletitle{Top-down versus bottom-up control of attention in
  the prefrontal and posterior parietal cortices}.
\newblock \bibinfo{journal}{\emph{Science}} \bibinfo{volume}{315},
  \bibinfo{number}{5820} (\bibinfo{year}{2007}), \bibinfo{pages}{1860--1862}.
\newblock


\bibitem[\protect\citeauthoryear{Cao, Ji, Cao, Chang, and Niebles}{Cao
  et~al\mbox{.}}{2020}]%
        {cao2020few}
\bibfield{author}{\bibinfo{person}{Kaidi Cao}, \bibinfo{person}{Jingwei Ji},
  \bibinfo{person}{Zhangjie Cao}, \bibinfo{person}{Chien-Yi Chang}, {and}
  \bibinfo{person}{Juan~Carlos Niebles}.} \bibinfo{year}{2020}\natexlab{}.
\newblock \showarticletitle{Few-shot video classification via temporal
  alignment}. In \bibinfo{booktitle}{\emph{Proceedings of the IEEE/CVF
  Conference on Computer Vision and Pattern Recognition (CVPR)}}.
  \bibinfo{pages}{10618--10627}.
\newblock


\bibitem[\protect\citeauthoryear{Carreira and Zisserman}{Carreira and
  Zisserman}{2017}]%
        {carreira2017quo}
\bibfield{author}{\bibinfo{person}{Joao Carreira} {and} \bibinfo{person}{Andrew
  Zisserman}.} \bibinfo{year}{2017}\natexlab{}.
\newblock \showarticletitle{Quo vadis, action recognition? a new model and the
  kinetics dataset}. In \bibinfo{booktitle}{\emph{Proceedings of the IEEE
  Conference on Computer Vision and Pattern Recognition (CVPR)}}.
  \bibinfo{pages}{6299--6308}.
\newblock


\bibitem[\protect\citeauthoryear{Chen, Kornblith, Norouzi, and Hinton}{Chen
  et~al\mbox{.}}{2020}]%
        {chen2020simple}
\bibfield{author}{\bibinfo{person}{Ting Chen}, \bibinfo{person}{Simon
  Kornblith}, \bibinfo{person}{Mohammad Norouzi}, {and}
  \bibinfo{person}{Geoffrey Hinton}.} \bibinfo{year}{2020}\natexlab{}.
\newblock \showarticletitle{A simple framework for contrastive learning of
  visual representations}. In \bibinfo{booktitle}{\emph{International
  Conference on Machine Learning (ICML)}}. PMLR, \bibinfo{pages}{1597--1607}.
\newblock


\bibitem[\protect\citeauthoryear{Chung, Wuu, Yang, Tai, and Tang}{Chung
  et~al\mbox{.}}{2020}]%
        {chung2020haa500}
\bibfield{author}{\bibinfo{person}{Jihoon Chung}, \bibinfo{person}{Cheng-hsin
  Wuu}, \bibinfo{person}{Hsuan-ru Yang}, \bibinfo{person}{Yu-Wing Tai}, {and}
  \bibinfo{person}{Chi-Keung Tang}.} \bibinfo{year}{2020}\natexlab{}.
\newblock \showarticletitle{HAA500: Human-centric atomic action dataset with
  curated videos}.
\newblock \bibinfo{journal}{\emph{arXiv preprint arXiv:2009.05224}}
  (\bibinfo{year}{2020}).
\newblock


\bibitem[\protect\citeauthoryear{Feichtenhofer, Fan, Malik, and
  He}{Feichtenhofer et~al\mbox{.}}{2019}]%
        {feichtenhofer2019slowfast}
\bibfield{author}{\bibinfo{person}{Christoph Feichtenhofer},
  \bibinfo{person}{Haoqi Fan}, \bibinfo{person}{Jitendra Malik}, {and}
  \bibinfo{person}{Kaiming He}.} \bibinfo{year}{2019}\natexlab{}.
\newblock \showarticletitle{Slowfast networks for video recognition}. In
  \bibinfo{booktitle}{\emph{Proceedings of the IEEE/CVF International
  Conference on Computer Vision (CVPR)}}. \bibinfo{pages}{6202--6211}.
\newblock


\bibitem[\protect\citeauthoryear{Finn, Abbeel, and Levine}{Finn
  et~al\mbox{.}}{2017}]%
        {finn2017model}
\bibfield{author}{\bibinfo{person}{Chelsea Finn}, \bibinfo{person}{Pieter
  Abbeel}, {and} \bibinfo{person}{Sergey Levine}.}
  \bibinfo{year}{2017}\natexlab{}.
\newblock \showarticletitle{Model-agnostic meta-learning for fast adaptation of
  deep networks}. In \bibinfo{booktitle}{\emph{International Conference on
  Machine Learning (ICML)}}. PMLR, \bibinfo{pages}{1126--1135}.
\newblock


\bibitem[\protect\citeauthoryear{Fu, Zhang, Wang, Fu, and Jiang}{Fu
  et~al\mbox{.}}{2020}]%
        {fu2020depth}
\bibfield{author}{\bibinfo{person}{Yuqian Fu}, \bibinfo{person}{Li Zhang},
  \bibinfo{person}{Junke Wang}, \bibinfo{person}{Yanwei Fu}, {and}
  \bibinfo{person}{Yu-Gang Jiang}.} \bibinfo{year}{2020}\natexlab{}.
\newblock \showarticletitle{Depth guided adaptive meta-fusion network for
  few-shot video recognition}. In \bibinfo{booktitle}{\emph{Proceedings of the
  ACM International Conference on Multimedia (MM)}}.
  \bibinfo{pages}{1142--1151}.
\newblock


\bibitem[\protect\citeauthoryear{Girdhar and Ramanan}{Girdhar and
  Ramanan}{2017}]%
        {girdhar2017attentional}
\bibfield{author}{\bibinfo{person}{Rohit Girdhar} {and} \bibinfo{person}{Deva
  Ramanan}.} \bibinfo{year}{2017}\natexlab{}.
\newblock \showarticletitle{Attentional pooling for action recognition}. In
  \bibinfo{booktitle}{\emph{Proceedings of the International Conference on
  Neural Information Processing Systems (NeurIPS)}}. \bibinfo{pages}{33--44}.
\newblock


\bibitem[\protect\citeauthoryear{Gutmann and Hyv{\"a}rinen}{Gutmann and
  Hyv{\"a}rinen}{2010}]%
        {gutmann2010noise}
\bibfield{author}{\bibinfo{person}{Michael Gutmann} {and} \bibinfo{person}{Aapo
  Hyv{\"a}rinen}.} \bibinfo{year}{2010}\natexlab{}.
\newblock \showarticletitle{Noise-contrastive estimation: A new estimation
  principle for unnormalized statistical models}. In
  \bibinfo{booktitle}{\emph{Proceedings of the International Conference on
  Artificial Intelligence and Statistics}}. JMLR Workshop and Conference
  Proceedings, \bibinfo{pages}{297--304}.
\newblock


\bibitem[\protect\citeauthoryear{He, Hong, Liu, Xu, Zha, and Wang}{He
  et~al\mbox{.}}{2020b}]%
        {he2020memory}
\bibfield{author}{\bibinfo{person}{Jun He}, \bibinfo{person}{Richang Hong},
  \bibinfo{person}{Xueliang Liu}, \bibinfo{person}{Mingliang Xu},
  \bibinfo{person}{Zheng-Jun Zha}, {and} \bibinfo{person}{Meng Wang}.}
  \bibinfo{year}{2020}\natexlab{b}.
\newblock \showarticletitle{Memory-augmented relation network for few-shot
  learning}. In \bibinfo{booktitle}{\emph{Proceedings of the ACM International
  Conference on Multimedia (MM)}}. \bibinfo{pages}{1236--1244}.
\newblock


\bibitem[\protect\citeauthoryear{He, Fan, Wu, Xie, and Girshick}{He
  et~al\mbox{.}}{2020a}]%
        {he2020momentum}
\bibfield{author}{\bibinfo{person}{Kaiming He}, \bibinfo{person}{Haoqi Fan},
  \bibinfo{person}{Yuxin Wu}, \bibinfo{person}{Saining Xie}, {and}
  \bibinfo{person}{Ross Girshick}.} \bibinfo{year}{2020}\natexlab{a}.
\newblock \showarticletitle{Momentum contrast for unsupervised visual
  representation learning}. In \bibinfo{booktitle}{\emph{Proceedings of the
  IEEE/CVF Conference on Computer Vision and Pattern Recognition (CVPR)}}.
  \bibinfo{pages}{9729--9738}.
\newblock


\bibitem[\protect\citeauthoryear{He, Zhang, Ren, and Sun}{He
  et~al\mbox{.}}{2016}]%
        {he2016deep}
\bibfield{author}{\bibinfo{person}{Kaiming He}, \bibinfo{person}{Xiangyu
  Zhang}, \bibinfo{person}{Shaoqing Ren}, {and} \bibinfo{person}{Jian Sun}.}
  \bibinfo{year}{2016}\natexlab{}.
\newblock \showarticletitle{Deep residual learning for image recognition}. In
  \bibinfo{booktitle}{\emph{Proceedings of the IEEE Conference on Computer
  Vision and Pattern Recognition (CVPR)}}. \bibinfo{pages}{770--778}.
\newblock


\bibitem[\protect\citeauthoryear{Khosla, Teterwak, Wang, Sarna, Tian, Isola,
  Maschinot, Liu, and Krishnan}{Khosla et~al\mbox{.}}{2020}]%
        {khosla2020supervised}
\bibfield{author}{\bibinfo{person}{Prannay Khosla}, \bibinfo{person}{Piotr
  Teterwak}, \bibinfo{person}{Chen Wang}, \bibinfo{person}{Aaron Sarna},
  \bibinfo{person}{Yonglong Tian}, \bibinfo{person}{Phillip Isola},
  \bibinfo{person}{Aaron Maschinot}, \bibinfo{person}{Ce Liu}, {and}
  \bibinfo{person}{Dilip Krishnan}.} \bibinfo{year}{2020}\natexlab{}.
\newblock \showarticletitle{Supervised contrastive learning}. In
  \bibinfo{booktitle}{\emph{Proceedings of the International Conference on
  Neural Information Processing Systems (NeurIPS)}}, Vol.~\bibinfo{volume}{33}.
\newblock


\bibitem[\protect\citeauthoryear{Kong, Huang, Qin, and Wang}{Kong
  et~al\mbox{.}}{2019}]%
        {kong2019joint}
\bibfield{author}{\bibinfo{person}{Longteng Kong}, \bibinfo{person}{Di Huang},
  \bibinfo{person}{Jie Qin}, {and} \bibinfo{person}{Yunhong Wang}.}
  \bibinfo{year}{2019}\natexlab{}.
\newblock \showarticletitle{A joint framework for athlete tracking and action
  recognition in sports videos}.
\newblock \bibinfo{journal}{\emph{IEEE Transactions on Circuits and Systems for
  Video Technology}} \bibinfo{volume}{30}, \bibinfo{number}{2}
  (\bibinfo{year}{2019}), \bibinfo{pages}{532--548}.
\newblock


\bibitem[\protect\citeauthoryear{Li, Chen, Li, and Yu}{Li
  et~al\mbox{.}}{2019}]%
        {li2019motion}
\bibfield{author}{\bibinfo{person}{Haofeng Li}, \bibinfo{person}{Guanqi Chen},
  \bibinfo{person}{Guanbin Li}, {and} \bibinfo{person}{Yizhou Yu}.}
  \bibinfo{year}{2019}\natexlab{}.
\newblock \showarticletitle{Motion guided attention for video salient object
  detection}. In \bibinfo{booktitle}{\emph{Proceedings of the IEEE/CVF
  International Conference on Computer Vision (ICCV)}}.
  \bibinfo{pages}{7274--7283}.
\newblock


\bibitem[\protect\citeauthoryear{Li, Zhou, Chen, and Li}{Li
  et~al\mbox{.}}{2017}]%
        {li2017meta}
\bibfield{author}{\bibinfo{person}{Zhenguo Li}, \bibinfo{person}{Fengwei Zhou},
  \bibinfo{person}{Fei Chen}, {and} \bibinfo{person}{Hang Li}.}
  \bibinfo{year}{2017}\natexlab{}.
\newblock \showarticletitle{Meta-sgd: Learning to learn quickly for few-shot
  learning}.
\newblock \bibinfo{journal}{\emph{arXiv preprint arXiv:1707.09835}}
  (\bibinfo{year}{2017}).
\newblock


\bibitem[\protect\citeauthoryear{Lin, Gan, and Han}{Lin et~al\mbox{.}}{2019}]%
        {lin2019tsm}
\bibfield{author}{\bibinfo{person}{Ji Lin}, \bibinfo{person}{Chuang Gan}, {and}
  \bibinfo{person}{Song Han}.} \bibinfo{year}{2019}\natexlab{}.
\newblock \showarticletitle{TSM: Temporal shift module for efficient video
  understanding}. In \bibinfo{booktitle}{\emph{Proceedings of the IEEE/CVF
  International Conference on Computer Vision (CVPR)}}.
  \bibinfo{pages}{7083--7093}.
\newblock


\bibitem[\protect\citeauthoryear{Navalpakkam and Itti}{Navalpakkam and
  Itti}{2006}]%
        {navalpakkam2006integrated}
\bibfield{author}{\bibinfo{person}{Vidhya Navalpakkam} {and}
  \bibinfo{person}{Laurent Itti}.} \bibinfo{year}{2006}\natexlab{}.
\newblock \showarticletitle{An integrated model of top-down and bottom-up
  attention for optimizing detection speed}. In
  \bibinfo{booktitle}{\emph{Proceedings of the IEEE Conference on Computer
  Vision and Pattern Recognition (CVPR)}}, Vol.~\bibinfo{volume}{2}. IEEE,
  \bibinfo{pages}{2049--2056}.
\newblock


\bibitem[\protect\citeauthoryear{Oliva, Torralba, Castelhano, and
  Henderson}{Oliva et~al\mbox{.}}{2003}]%
        {oliva2003top}
\bibfield{author}{\bibinfo{person}{Aude Oliva}, \bibinfo{person}{Antonio
  Torralba}, \bibinfo{person}{Monica~S Castelhano}, {and}
  \bibinfo{person}{John~M Henderson}.} \bibinfo{year}{2003}\natexlab{}.
\newblock \showarticletitle{Top-down control of visual attention in object
  detection}. In \bibinfo{booktitle}{\emph{Proceedings of the International
  Conference on Image Processing (ICIP)}}, Vol.~\bibinfo{volume}{1}. IEEE,
  \bibinfo{pages}{I--253}.
\newblock


\bibitem[\protect\citeauthoryear{Oord, Li, and Vinyals}{Oord
  et~al\mbox{.}}{2018}]%
        {oord2018representation}
\bibfield{author}{\bibinfo{person}{Aaron van~den Oord}, \bibinfo{person}{Yazhe
  Li}, {and} \bibinfo{person}{Oriol Vinyals}.} \bibinfo{year}{2018}\natexlab{}.
\newblock \showarticletitle{Representation learning with contrastive predictive
  coding}.
\newblock \bibinfo{journal}{\emph{arXiv preprint arXiv:1807.03748}}
  (\bibinfo{year}{2018}).
\newblock


\bibitem[\protect\citeauthoryear{Perrett, Masullo, Burghardt, Mirmehdi, and
  Damen}{Perrett et~al\mbox{.}}{2021}]%
        {perrett2021temporal}
\bibfield{author}{\bibinfo{person}{Toby Perrett}, \bibinfo{person}{Alessandro
  Masullo}, \bibinfo{person}{Tilo Burghardt}, \bibinfo{person}{Majid Mirmehdi},
  {and} \bibinfo{person}{Dima Damen}.} \bibinfo{year}{2021}\natexlab{}.
\newblock \showarticletitle{Temporal-relational crossTransformers for few-shot
  action recognition}. In \bibinfo{booktitle}{\emph{Proceedings of the IEEE/CVF
  Conference on Computer Vision and Pattern Recognition (CVPR)}}.
  \bibinfo{pages}{475--484}.
\newblock


\bibitem[\protect\citeauthoryear{Piergiovanni and Ryoo}{Piergiovanni and
  Ryoo}{2018}]%
        {piergiovanni2018fine}
\bibfield{author}{\bibinfo{person}{AJ Piergiovanni} {and}
  \bibinfo{person}{Michael~S Ryoo}.} \bibinfo{year}{2018}\natexlab{}.
\newblock \showarticletitle{Fine-grained activity recognition in baseball
  videos}. In \bibinfo{booktitle}{\emph{Proceedings of the IEEE Conference on
  Computer Vision and Pattern Recognition Workshops (CVPRW)}}.
  \bibinfo{pages}{1740--1748}.
\newblock


\bibitem[\protect\citeauthoryear{Posner and Petersen}{Posner and
  Petersen}{1990}]%
        {posner1990attention}
\bibfield{author}{\bibinfo{person}{Michael~I Posner} {and}
  \bibinfo{person}{Steven~E Petersen}.} \bibinfo{year}{1990}\natexlab{}.
\newblock \showarticletitle{The attention system of the human brain}.
\newblock \bibinfo{journal}{\emph{Annual Review of Neuroscience}}
  \bibinfo{volume}{13}, \bibinfo{number}{1} (\bibinfo{year}{1990}),
  \bibinfo{pages}{25--42}.
\newblock


\bibitem[\protect\citeauthoryear{Qi, Qin, Yang, Wang, and Luo}{Qi
  et~al\mbox{.}}{2021}]%
        {qi2021semantics}
\bibfield{author}{\bibinfo{person}{Mengshi Qi}, \bibinfo{person}{Jie Qin},
  \bibinfo{person}{Yi Yang}, \bibinfo{person}{Yunhong Wang}, {and}
  \bibinfo{person}{Jiebo Luo}.} \bibinfo{year}{2021}\natexlab{}.
\newblock \showarticletitle{Semantics-aware spatial-temporal binaries for
  cross-modal video retrieval}.
\newblock \bibinfo{journal}{\emph{IEEE Transactions on Image Processing}}
  \bibinfo{volume}{30} (\bibinfo{year}{2021}), \bibinfo{pages}{2989--3004}.
\newblock


\bibitem[\protect\citeauthoryear{Qi, Qin, Zhen, Huang, Yang, and Luo}{Qi
  et~al\mbox{.}}{2020a}]%
        {qi2020few}
\bibfield{author}{\bibinfo{person}{Mengshi Qi}, \bibinfo{person}{Jie Qin},
  \bibinfo{person}{Xiantong Zhen}, \bibinfo{person}{Di Huang},
  \bibinfo{person}{Yi Yang}, {and} \bibinfo{person}{Jiebo Luo}.}
  \bibinfo{year}{2020}\natexlab{a}.
\newblock \showarticletitle{Few-shot ensemble learning for video classification
  with slowFast memory networks}. In \bibinfo{booktitle}{\emph{Proceedings of
  the ACM International Conference on Multimedia (MM)}}.
  \bibinfo{pages}{3007--3015}.
\newblock


\bibitem[\protect\citeauthoryear{Qi, Wang, Li, and Luo}{Qi
  et~al\mbox{.}}{2019a}]%
        {qi2019sports}
\bibfield{author}{\bibinfo{person}{Mengshi Qi}, \bibinfo{person}{Yunhong Wang},
  \bibinfo{person}{Annan Li}, {and} \bibinfo{person}{Jiebo Luo}.}
  \bibinfo{year}{2019}\natexlab{a}.
\newblock \showarticletitle{Sports video captioning via attentive motion
  representation and group relationship modeling}.
\newblock \bibinfo{journal}{\emph{IEEE Transactions on Circuits and Systems for
  Video Technology}} \bibinfo{volume}{30}, \bibinfo{number}{8}
  (\bibinfo{year}{2019}), \bibinfo{pages}{2617--2633}.
\newblock


\bibitem[\protect\citeauthoryear{Qi, Wang, Li, and Luo}{Qi
  et~al\mbox{.}}{2020b}]%
        {qi2020stc}
\bibfield{author}{\bibinfo{person}{Mengshi Qi}, \bibinfo{person}{Yunhong Wang},
  \bibinfo{person}{Annan Li}, {and} \bibinfo{person}{Jiebo Luo}.}
  \bibinfo{year}{2020}\natexlab{b}.
\newblock \showarticletitle{STC-GAN: Spatio-temporally coupled generative
  adversarial networks for predictive scene parsing}.
\newblock \bibinfo{journal}{\emph{IEEE Transactions on Image Processing}}
  \bibinfo{volume}{29} (\bibinfo{year}{2020}), \bibinfo{pages}{5420--5430}.
\newblock


\bibitem[\protect\citeauthoryear{Qi, Wang, Qin, Li, Luo, and Van~Gool}{Qi
  et~al\mbox{.}}{2019b}]%
        {qi2019stagnet}
\bibfield{author}{\bibinfo{person}{Mengshi Qi}, \bibinfo{person}{Yunhong Wang},
  \bibinfo{person}{Jie Qin}, \bibinfo{person}{Annan Li}, \bibinfo{person}{Jiebo
  Luo}, {and} \bibinfo{person}{Luc Van~Gool}.}
  \bibinfo{year}{2019}\natexlab{b}.
\newblock \showarticletitle{stagNet: An attentive semantic RNN for group
  activity and individual action recognition}.
\newblock \bibinfo{journal}{\emph{IEEE Transactions on Circuits and Systems for
  Video Technology}} \bibinfo{volume}{30}, \bibinfo{number}{2}
  (\bibinfo{year}{2019}), \bibinfo{pages}{549--565}.
\newblock


\bibitem[\protect\citeauthoryear{Rohrbach, Rohrbach, Regneri, Amin, Andriluka,
  Pinkal, and Schiele}{Rohrbach et~al\mbox{.}}{2016}]%
        {rohrbach2016recognizing}
\bibfield{author}{\bibinfo{person}{Marcus Rohrbach}, \bibinfo{person}{Anna
  Rohrbach}, \bibinfo{person}{Michaela Regneri}, \bibinfo{person}{Sikandar
  Amin}, \bibinfo{person}{Mykhaylo Andriluka}, \bibinfo{person}{Manfred
  Pinkal}, {and} \bibinfo{person}{Bernt Schiele}.}
  \bibinfo{year}{2016}\natexlab{}.
\newblock \showarticletitle{Recognizing fine-grained and composite activities
  using hand-centric features and script data}.
\newblock \bibinfo{journal}{\emph{International Journal of Computer Vision}}
  \bibinfo{volume}{119}, \bibinfo{number}{3} (\bibinfo{year}{2016}),
  \bibinfo{pages}{346--373}.
\newblock


\bibitem[\protect\citeauthoryear{Rutishauser, Walther, Koch, and
  Perona}{Rutishauser et~al\mbox{.}}{2004}]%
        {rutishauser2004bottom}
\bibfield{author}{\bibinfo{person}{Ueli Rutishauser}, \bibinfo{person}{Dirk
  Walther}, \bibinfo{person}{Christof Koch}, {and} \bibinfo{person}{Pietro
  Perona}.} \bibinfo{year}{2004}\natexlab{}.
\newblock \showarticletitle{Is bottom-up attention useful for object
  recognition?}. In \bibinfo{booktitle}{\emph{Proceedings of the IEEE
  Conference on Computer Vision and Pattern Recognition (CVPR)}},
  Vol.~\bibinfo{volume}{2}. IEEE, \bibinfo{pages}{II--II}.
\newblock


\bibitem[\protect\citeauthoryear{Shao, Zhao, Dai, and Lin}{Shao
  et~al\mbox{.}}{2020}]%
        {shao2020finegym}
\bibfield{author}{\bibinfo{person}{Dian Shao}, \bibinfo{person}{Yue Zhao},
  \bibinfo{person}{Bo Dai}, {and} \bibinfo{person}{Dahua Lin}.}
  \bibinfo{year}{2020}\natexlab{}.
\newblock \showarticletitle{Finegym: A hierarchical video dataset for
  fine-grained action understanding}. In \bibinfo{booktitle}{\emph{Proceedings
  of the IEEE/CVF Conference on Computer Vision and Pattern Recognition
  (CVPR)}}. \bibinfo{pages}{2616--2625}.
\newblock


\bibitem[\protect\citeauthoryear{Snell, Swersky, and Zemel}{Snell
  et~al\mbox{.}}{2017}]%
        {snell2017prototypical}
\bibfield{author}{\bibinfo{person}{Jake Snell}, \bibinfo{person}{Kevin
  Swersky}, {and} \bibinfo{person}{Richard Zemel}.}
  \bibinfo{year}{2017}\natexlab{}.
\newblock \showarticletitle{Prototypical networks for few-shot learning}. In
  \bibinfo{booktitle}{\emph{Proceedings of the International Conference on
  Neural Information Processing Systems (NeurIPS)}}.
  \bibinfo{pages}{4080--4090}.
\newblock


\bibitem[\protect\citeauthoryear{Soomro, Zamir, and Shah}{Soomro
  et~al\mbox{.}}{2012}]%
        {soomro2012ucf101}
\bibfield{author}{\bibinfo{person}{Khurram Soomro},
  \bibinfo{person}{Amir~Roshan Zamir}, {and} \bibinfo{person}{Mubarak Shah}.}
  \bibinfo{year}{2012}\natexlab{}.
\newblock \showarticletitle{UCF101: A dataset of 101 human actions classes from
  videos in the wild}.
\newblock \bibinfo{journal}{\emph{arXiv preprint arXiv:1212.0402}}
  (\bibinfo{year}{2012}).
\newblock


\bibitem[\protect\citeauthoryear{Sun, Wang, Liang, and He}{Sun
  et~al\mbox{.}}{2017}]%
        {sun2017taichi}
\bibfield{author}{\bibinfo{person}{Shan Sun}, \bibinfo{person}{Feng Wang},
  \bibinfo{person}{Qi Liang}, {and} \bibinfo{person}{Liang He}.}
  \bibinfo{year}{2017}\natexlab{}.
\newblock \showarticletitle{Taichi: A fine-grained action recognition dataset}.
  In \bibinfo{booktitle}{\emph{Proceedings of the ACM International Conference
  on Multimedia Retrieval (ICMR)}}. \bibinfo{pages}{429--433}.
\newblock


\bibitem[\protect\citeauthoryear{Sung, Yang, Zhang, Xiang, Torr, and
  Hospedales}{Sung et~al\mbox{.}}{2018}]%
        {sung2018learning}
\bibfield{author}{\bibinfo{person}{Flood Sung}, \bibinfo{person}{Yongxin Yang},
  \bibinfo{person}{Li Zhang}, \bibinfo{person}{Tao Xiang},
  \bibinfo{person}{Philip~HS Torr}, {and} \bibinfo{person}{Timothy~M
  Hospedales}.} \bibinfo{year}{2018}\natexlab{}.
\newblock \showarticletitle{Learning to compare: Relation network for few-shot
  learning}. In \bibinfo{booktitle}{\emph{Proceedings of the IEEE Conference on
  Computer Vision and Pattern Recognition (CVPR)}}.
  \bibinfo{pages}{1199--1208}.
\newblock


\bibitem[\protect\citeauthoryear{Ungerleider and G}{Ungerleider and G}{2000}]%
        {ungerleider2000mechanisms}
\bibfield{author}{\bibinfo{person}{Sabine~Kastner Ungerleider} {and}
  \bibinfo{person}{Leslie G}.} \bibinfo{year}{2000}\natexlab{}.
\newblock \showarticletitle{Mechanisms of visual attention in the human
  cortex}.
\newblock \bibinfo{journal}{\emph{Annual Review of Neuroscience}}
  \bibinfo{volume}{23}, \bibinfo{number}{1} (\bibinfo{year}{2000}),
  \bibinfo{pages}{315--341}.
\newblock


\bibitem[\protect\citeauthoryear{Van~der Maaten and Hinton}{Van~der Maaten and
  Hinton}{2008}]%
        {van2008visualizing}
\bibfield{author}{\bibinfo{person}{Laurens Van~der Maaten} {and}
  \bibinfo{person}{Geoffrey Hinton}.} \bibinfo{year}{2008}\natexlab{}.
\newblock \showarticletitle{Visualizing data using t-SNE.}
\newblock \bibinfo{journal}{\emph{Journal of Machine Learning Research}}
  \bibinfo{volume}{9}, \bibinfo{number}{11} (\bibinfo{year}{2008}).
\newblock


\bibitem[\protect\citeauthoryear{Vinyals, Blundell, Lillicrap, Kavukcuoglu, and
  Wierstra}{Vinyals et~al\mbox{.}}{2016}]%
        {vinyals2016matching}
\bibfield{author}{\bibinfo{person}{Oriol Vinyals}, \bibinfo{person}{Charles
  Blundell}, \bibinfo{person}{Timothy Lillicrap}, \bibinfo{person}{Koray
  Kavukcuoglu}, {and} \bibinfo{person}{Daan Wierstra}.}
  \bibinfo{year}{2016}\natexlab{}.
\newblock \showarticletitle{Matching networks for one shot learning}. In
  \bibinfo{booktitle}{\emph{Proceedings of the International Conference on
  Neural Information Processing Systems (NeurIPS)}}.
  \bibinfo{pages}{3637--3645}.
\newblock


\bibitem[\protect\citeauthoryear{Wang, Jiang, Qian, Yang, Li, Zhang, Wang, and
  Tang}{Wang et~al\mbox{.}}{2017}]%
        {wang2017residual}
\bibfield{author}{\bibinfo{person}{Fei Wang}, \bibinfo{person}{Mengqing Jiang},
  \bibinfo{person}{Chen Qian}, \bibinfo{person}{Shuo Yang},
  \bibinfo{person}{Cheng Li}, \bibinfo{person}{Honggang Zhang},
  \bibinfo{person}{Xiaogang Wang}, {and} \bibinfo{person}{Xiaoou Tang}.}
  \bibinfo{year}{2017}\natexlab{}.
\newblock \showarticletitle{Residual attention network for image
  classification}. In \bibinfo{booktitle}{\emph{Proceedings of the IEEE
  Conference on Computer Vision and Pattern Recognition (CVPR)}}.
  \bibinfo{pages}{3156--3164}.
\newblock


\bibitem[\protect\citeauthoryear{Wang, Du, Li, and Wang}{Wang
  et~al\mbox{.}}{2019}]%
        {wang2019atrous}
\bibfield{author}{\bibinfo{person}{Jiahao Wang}, \bibinfo{person}{Zhengyin Du},
  \bibinfo{person}{Annan Li}, {and} \bibinfo{person}{Yunhong Wang}.}
  \bibinfo{year}{2019}\natexlab{}.
\newblock \showarticletitle{Atrous temporal convolutional network for video
  action segmentation}. In \bibinfo{booktitle}{\emph{Proceedings of the
  International Conference on Image Processing (ICIP)}}. IEEE,
  \bibinfo{pages}{1585--1589}.
\newblock


\bibitem[\protect\citeauthoryear{Wang, Xiong, Wang, Qiao, Lin, Tang, and
  Van~Gool}{Wang et~al\mbox{.}}{2016}]%
        {wang2016temporal}
\bibfield{author}{\bibinfo{person}{Limin Wang}, \bibinfo{person}{Yuanjun
  Xiong}, \bibinfo{person}{Zhe Wang}, \bibinfo{person}{Yu Qiao},
  \bibinfo{person}{Dahua Lin}, \bibinfo{person}{Xiaoou Tang}, {and}
  \bibinfo{person}{Luc Van~Gool}.} \bibinfo{year}{2016}\natexlab{}.
\newblock \showarticletitle{Temporal segment networks: Towards good practices
  for deep action recognition}. In \bibinfo{booktitle}{\emph{Proceedings of the
  European Conference on Computer Vision (ECCV)}}. Springer,
  \bibinfo{pages}{20--36}.
\newblock


\bibitem[\protect\citeauthoryear{Wang, Zhao, Li, and Tian}{Wang
  et~al\mbox{.}}{2020}]%
        {wang2020cooperative}
\bibfield{author}{\bibinfo{person}{Zeyuan Wang}, \bibinfo{person}{Yifan Zhao},
  \bibinfo{person}{Jia Li}, {and} \bibinfo{person}{Yonghong Tian}.}
  \bibinfo{year}{2020}\natexlab{}.
\newblock \showarticletitle{Cooperative bi-path metric for few-shot learning}.
  In \bibinfo{booktitle}{\emph{Proceedings of the ACM International Conference
  on Multimedia (MM)}}. \bibinfo{pages}{1524--1532}.
\newblock


\bibitem[\protect\citeauthoryear{Woo, Park, Lee, and Kweon}{Woo
  et~al\mbox{.}}{2018}]%
        {woo2018cbam}
\bibfield{author}{\bibinfo{person}{Sanghyun Woo}, \bibinfo{person}{Jongchan
  Park}, \bibinfo{person}{Joon-Young Lee}, {and} \bibinfo{person}{In~So
  Kweon}.} \bibinfo{year}{2018}\natexlab{}.
\newblock \showarticletitle{CBAM: Convolutional block attention module}. In
  \bibinfo{booktitle}{\emph{Proceedings of the European Conference on Computer
  Vision (ECCV)}}. \bibinfo{pages}{3--19}.
\newblock


\bibitem[\protect\citeauthoryear{Yan, Ni, and Yang}{Yan et~al\mbox{.}}{2017}]%
        {yan2017fine}
\bibfield{author}{\bibinfo{person}{Yichao Yan}, \bibinfo{person}{Bingbing Ni},
  {and} \bibinfo{person}{Xiaokang Yang}.} \bibinfo{year}{2017}\natexlab{}.
\newblock \showarticletitle{Fine-grained recognition via attribute-guided
  attentive feature aggregation}. In \bibinfo{booktitle}{\emph{Proceedings of
  the ACM International Conference on Multimedia (MM)}}.
  \bibinfo{pages}{1032--1040}.
\newblock


\bibitem[\protect\citeauthoryear{Yang, Xu, Shi, Dai, and Zhou}{Yang
  et~al\mbox{.}}{2020}]%
        {yang2020temporal}
\bibfield{author}{\bibinfo{person}{Ceyuan Yang}, \bibinfo{person}{Yinghao Xu},
  \bibinfo{person}{Jianping Shi}, \bibinfo{person}{Bo Dai}, {and}
  \bibinfo{person}{Bolei Zhou}.} \bibinfo{year}{2020}\natexlab{}.
\newblock \showarticletitle{Temporal pyramid network for action recognition}.
  In \bibinfo{booktitle}{\emph{Proceedings of the IEEE/CVF Conference on
  Computer Vision and Pattern Recognition (CVPR)}}. \bibinfo{pages}{591--600}.
\newblock


\bibitem[\protect\citeauthoryear{Zhang, Zhang, Qi, Li, Torr, and Koniusz}{Zhang
  et~al\mbox{.}}{2020b}]%
        {zhang2020few}
\bibfield{author}{\bibinfo{person}{Hongguang Zhang}, \bibinfo{person}{Li
  Zhang}, \bibinfo{person}{Xiaojuan Qi}, \bibinfo{person}{Hongdong Li},
  \bibinfo{person}{Philip~HS Torr}, {and} \bibinfo{person}{Piotr Koniusz}.}
  \bibinfo{year}{2020}\natexlab{b}.
\newblock \showarticletitle{Few-shot action recognition with
  permutation-invariant attention}. In \bibinfo{booktitle}{\emph{Proceedings of
  the European Conference on Computer Vision (ECCV)}}. Springer.
\newblock


\bibitem[\protect\citeauthoryear{Zhang, Chang, Liu, Luo, Prakash, and
  Hauptmann}{Zhang et~al\mbox{.}}{2020a}]%
        {zhang2020few-shot}
\bibfield{author}{\bibinfo{person}{Lingling Zhang}, \bibinfo{person}{Xiaojun
  Chang}, \bibinfo{person}{Jun Liu}, \bibinfo{person}{Minnan Luo},
  \bibinfo{person}{Mahesh Prakash}, {and} \bibinfo{person}{Alexander~G
  Hauptmann}.} \bibinfo{year}{2020}\natexlab{a}.
\newblock \showarticletitle{Few-shot activity recognition with cross-modal
  memory network}.
\newblock \bibinfo{journal}{\emph{Pattern Recognition}}  \bibinfo{volume}{108}
  (\bibinfo{year}{2020}), \bibinfo{pages}{107348}.
\newblock


\bibitem[\protect\citeauthoryear{Zhu, Tan, Zhou, Liu, Yue, Ding, and Ma}{Zhu
  et~al\mbox{.}}{2018}]%
        {zhu2018fine}
\bibfield{author}{\bibinfo{person}{Chen Zhu}, \bibinfo{person}{Xiao Tan},
  \bibinfo{person}{Feng Zhou}, \bibinfo{person}{Xiao Liu},
  \bibinfo{person}{Kaiyu Yue}, \bibinfo{person}{Errui Ding}, {and}
  \bibinfo{person}{Yi Ma}.} \bibinfo{year}{2018}\natexlab{}.
\newblock \showarticletitle{Fine-grained video categorization with redundancy
  reduction attention}. In \bibinfo{booktitle}{\emph{Proceedings of the
  European Conference on Computer Vision (ECCV)}}. \bibinfo{pages}{136--152}.
\newblock


\bibitem[\protect\citeauthoryear{Zhu and Yang}{Zhu and Yang}{2018}]%
        {zhu2018compound}
\bibfield{author}{\bibinfo{person}{Linchao Zhu} {and} \bibinfo{person}{Yi
  Yang}.} \bibinfo{year}{2018}\natexlab{}.
\newblock \showarticletitle{Compound memory networks for few-shot video
  classification}. In \bibinfo{booktitle}{\emph{Proceedings of the European
  Conference on Computer Vision (ECCV)}}. \bibinfo{pages}{751--766}.
\newblock


\bibitem[\protect\citeauthoryear{Zhu, Toisoul, Prez-Ra, Zhang, Martinez, and
  Xiang}{Zhu et~al\mbox{.}}{2021}]%
        {zhu2021few}
\bibfield{author}{\bibinfo{person}{Xiatian Zhu}, \bibinfo{person}{Antoine
  Toisoul}, \bibinfo{person}{Juan-Manuel Prez-Ra}, \bibinfo{person}{Li Zhang},
  \bibinfo{person}{Brais Martinez}, {and} \bibinfo{person}{Tao Xiang}.}
  \bibinfo{year}{2021}\natexlab{}.
\newblock \showarticletitle{Few-shot action recognition with prototype-centered
  attentive learning}.
\newblock \bibinfo{journal}{\emph{arXiv preprint arXiv:2101.08085}}
  (\bibinfo{year}{2021}).
\newblock


\bibitem[\protect\citeauthoryear{Zou, Zhang, Chen, Tian, Wang, and Moura}{Zou
  et~al\mbox{.}}{2020}]%
        {zou2020compositional}
\bibfield{author}{\bibinfo{person}{Yixiong Zou}, \bibinfo{person}{Shanghang
  Zhang}, \bibinfo{person}{Ke Chen}, \bibinfo{person}{Yonghong Tian},
  \bibinfo{person}{Yaowei Wang}, {and} \bibinfo{person}{Jos{\'e}~MF Moura}.}
  \bibinfo{year}{2020}\natexlab{}.
\newblock \showarticletitle{Compositional few-shot recognition with primitive
  discovery and enhancing}. In \bibinfo{booktitle}{\emph{Proceedings of the ACM
  International Conference on Multimedia (MM)}}. \bibinfo{pages}{156--164}.
\newblock


\end{thebibliography}


\end{document}